\pdfoutput=1

\documentclass[11pt]{article}

\usepackage[]{ACL2023}

\usepackage{times}
\usepackage{latexsym}

\usepackage[T1]{fontenc}

\usepackage[utf8]{inputenc}

\usepackage{stmaryrd}
\usepackage{microtype}

\usepackage{inconsolata}

\usepackage{amsmath, amsmath}
\usepackage{makecell}
\usepackage{multirow}
\usepackage{array}
\usepackage{booktabs}
\usepackage{soul}
\usepackage[normalem]{ulem}
\usepackage{listings}
\usepackage[switch]{lineno}  %
\usepackage{enumitem}
\usepackage{amssymb}
\usepackage{bbm}
\usepackage[linesnumbered,ruled,vlined]{algorithm2e}

\usepackage{color,xcolor,colortbl}
\usepackage{mathrsfs}
\usepackage{graphicx, float, subfigure}

\DeclareMathOperator*{\argmin}{arg\,min}

\DeclareMathOperator*{\natlog}{natlog}

\newcommand{\ie}{\textit{i}.\textit{e}., }

\title{\texttt{NatLogAttack}: A Framework for Attacking Natural Language Inference Models with Natural Logic}

\author{
Zi'ou Zheng \&~\textbf{Xiaodan Zhu} \\ 
  Department of Electrical and Computer Engineering
  \& Ingenuity Labs Research Institute \\
  Queen's University \\ 
  \texttt{\{ziou.zheng,xiaodan.zhu\}@queensu.ca} \\
  }

\begin{document}
\maketitle
\begin{abstract}
Reasoning has been a central topic in artificial intelligence from the beginning.
The recent progress made on distributed representation and neural networks continues to improve the state-of-the-art performance of natural language inference. However, it remains an open question whether the models perform real reasoning to reach their conclusions or rely on spurious correlations. 
Adversarial attacks have proven to be an important tool to help evaluate the Achilles' heel of the victim models. In this study, we explore the fundamental problem of developing attack models based on logic formalism. We propose \texttt{NatLogAttack} to perform systematic attacks centring around \textit{natural logic}, a classical logic formalism that is traceable back to Aristotle's syllogism and has been closely developed for natural language inference. The proposed framework renders both label-preserving and label-flipping attacks.
We show that compared to the existing attack models, \texttt{NatLogAttack} generates better adversarial examples with fewer visits to the victim models. 
The victim models are found to be more vulnerable under the label-flipping setting. 
\texttt{NatLogAttack} provides a tool to probe the existing and future NLI models' capacity from a key viewpoint and
we hope more logic-based attacks will be further explored for understanding the desired property of reasoning.~\footnote{The code of \texttt{NatLogAttack}  is available at  \url{https://github.com/orianna-zzo/NatLogAttack}.}


\end{abstract}

\section{Introduction}
%
While deep neural networks have achieved the state-of-the-art performance on a wide range of tasks, the models are often vulnerable and easily deceived by imposing perturbations to the original input~\cite{goodfellow2014explaining,kurakin2016adversarial}, which seriously hurts the accountability of the systems. In depth, this pertains to model robustness, capacity, and the development of models with more advanced intelligence.  

Natural language inference (NLI), also known as textual entailment~\cite{DBLP:conf/mlcw/DaganGM05,Iftene:W07-1421, maccartney2009natural,snli}, is a fundamental problem that models the inferential relationships between a premise and hypothesis sentence. The models built on \textit{distributed} representation have significantly improved the performance on different benchmarks \cite{snli,Qian2017esim,mnli,kim2018,bert,roberta,zhang2020semantics, pilault2021conditionally}. However, it is still highly desirable to conduct research to probe if the models possess the desired reasoning ability rather than rely on spurious correlation to reach their conclusions~\cite{breakingNLI,poliak2018collecting,belinkov2019don,mccoy2019right,fragments2020}. 

Adversarial attacks have proven to be an important tool to reveal the Achilles' heel of victim models. Specifically for natural language inference, the logic relations are easily broken if an attack model does not properly generate the adversarial examples following the logic relations and related semantics. Therefore, unlike other textual attack tasks such as those relying on semantic similarity and relatedness, it is more challenging to create effective attacks here. 



In this study, we explore the basic problem of developing adversarial attacks based on logic formalism, with the aim to probe victim models for the desired reasoning capability. 
Specifically, we propose \texttt{NatLogAttack}, in which the adversarial attacks are generated based on \textit{natural logic} ~\cite{lakoff1970linguistics,van1995language,maccartney2009natural,icard2012inclusion,angeli2016combining,hu2018polarity,chen2021neurallog}, a classical logic formalism with a long history that has been closely developed with natural language inference. From a general perspective, natural language inference provides an appropriate setup for probing the development of \textit{distributed representation} and the models based on that. A robust solution for the task requires manipulation of discrete operations and adversarial attacks can help understand whether and how the required symbols and inference steps emerge from the data and the learned distributed representation. Our work has also been inspired by recent research on exploring the complementary strengths of neural networks and symbolic models~\cite{garcez2015neural,yang2017differentiable,rocktaschel2017end,evans2018learning,weber2019nlprolog,de2019neuro,Mao2019NeuroSymbolic,feng2020exploring,Feng2022}. 



Our research contributes to the development of logic-based adversarial attacks for natural language understanding. Specifically, we propose a novel attack framework, \texttt{NatLogAttack}, based on natural logic for natural language inference. Our experiments with both human and automatic evaluation show that the proposed model outperforms the state-of-the-art attack methods. Compared to the existing attack models, \texttt{NatLogAttack} generates better adversarial examples with fewer visits to the victim models. In addition to the commonly used attack setting where the labels of generated examples remain the same as the original pairs, we also propose to construct label-flipping attacks. The victim models are found to be more vulnerable in this setup and \texttt{NatLogAttack}
succeeds in deceiving them with much smaller
numbers of queries. 
\texttt{NatLogAttack} provides a systematic  approach to probing the existing and future NLI models' capacity from a basic viewpoint that has a traceable history, by combining it with the recent development of attacking models. The proposed framework is constrained by the natural logic formalism and we hope more logic-based attacks will be
further explored for understanding the desired property of natural language reasoning.

\section{Related Work}
\paragraph{Adversarial Attacks in NLP.}


White-box attacks leverage the architecture and parameters of victim models to craft adversarial examples~\cite{liang2017deep,wallace2019universal,ebrahimi2017hotflip}. Black-box models, however, have no such knowledge. Pioneering blind models ~\cite{jia2017adversarial}, for example, create adversarial examples by adding distracting sentences to the input. More recently, score-based (e.g., ~\citet{zhang2020generating,textfooler}) and decision-based attack models~\cite{zhao2017generating} also query the prediction scores or the final decisions of victim models. 

In terms of perturbation granularities, character-level attacks modify characters~\cite{ebrahimi2017hotflip} while word-level models rely on word substitutions that can be performed based on word embeddings~\cite{sato2018interpretable}, language models~\cite{zhang2020generating}, or even external knowledge bases~\cite{pso}. 
Sentence-level attack models add perturbation to an entire sentence by performing paraphrasing~\cite{SCPN} or attaching distracting sentences~\cite{jia2017adversarial}.


\citet{kang18acl} generated natural language inference examples based on entailment label composition functions with the help of lexical knowledge. \citet{minervini2018adversarially} utilized a set of first-order-logic constraints to measure the degree of rule violation for natural language inference. The efforts utilized the generated examples for data augmentation. The focus is not on adversarial attack and the adversarial examples' quality, e.g., the attack validity, is not evaluated. 

\paragraph{Natural Logic.} 
Natural logic has a long history and has been closely developed with natural language inference~\cite{lakoff1970linguistics,van1995language,maccartney2009natural,icard2012inclusion,angeli2016combining,hu2018polarity,chen2021neurallog}.
Recently, some efforts have started to consider monotonicity in attacks, including creating test 
sets to understand NLI models' behaviour~\cite{fragments2020,med2019,help2019,systematicity2020,geiger2020neural}. 
The existing work, however, has not performed systematic attacks based on natural logic. The core idea of monotonicity (e.g., downward monotone) and projection has not been systematically considered. The models have not been combined with the state-of-the-art adversarial attack framework and search strategies for the general purpose of adversarial attacks. For example, \citet{fragments2020} and \citet{systematicity2020} generate adversarial examples from a small vocabulary and pre-designed sentence structures. The effort of \citet{help2019} is limited by only considering one-edit distance  between a premise and hypothesis. We aim to explore principled approaches to constructing perturbations based on natural logic, and the control of the quality of attack generation can leverage the continuing advancement of language models. The proposed attack settings, along with the breakdown of attack categories, help reveal the properties of victim models in both label-preserving and label-flipping attacks.

\begin{figure*}[t]
\centering
\begin{minipage} [t]{1\linewidth}
\centering
    \includegraphics[width=1\linewidth,clip]{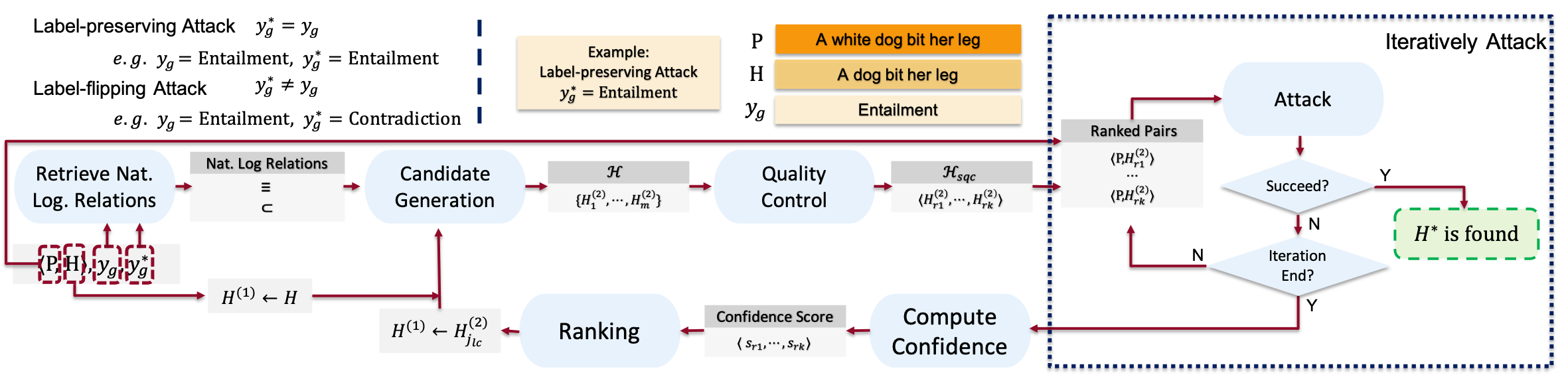}
\end{minipage}
\caption{Overview of \texttt{NatLogAttack} generation and attacking process.}
\label{fig:overview}
\vspace{-1em}
\end{figure*}

\section{\texttt{NatLogAttack}: A Natural-logic-based Attack Framework}

This section introduces \texttt{NatLogAttack}, a systematic adversarial attack framework centring around natural logic. The overview of \texttt{NatLogAttack}'s generation and attack process is depicted in Figure~\ref{fig:overview}. Below we will introduce the background, attack principles, setups, and each component of the framework.

\subsection{Background}
\label{sec:mainbackground}
The study of natural logic 
can be traced back to Aristotle's  syllogisms. Rather than performing deduction over an
abstract logical form, natural logic models inference in natural language by operating on the structure or surface form of language~\cite{lakoff1970linguistics,van1988semantics, valencia1991studies, van1995language, nairn2006computing, maccartney2009natural, maccartney2009extend, icard2012inclusion, angeli2014naturalli,hu2018polarity,chen2021monotonicity,chen2021neurallog}. It allows for a wide range of intuitive inferences in a conceptually clean way that we use daily and provides a good framework for attacking inference models---we doubt that a victim model vulnerable to such natural attacks indeed performs reliable reasoning. Our work uses the natural logic variant proposed by \citet{maccartney2009extend} and \citet{maccartney2009natural}, which extends the prior formalism to model the entailment relations between two spans of texts with seven relations $\frak{B}=\{\,\equiv, \sqsubset, \sqsupset, \wedge, \,\mid\,,  \smallsmile,\allowbreak \#\,\}$, representing \textit{equivalence}, \textit{forward entailment}, \textit{reverse entailment}, \textit{negation}, \textit{alternation}, \textit{cover}, and \textit{independence}, respectively. Through projection based on \textit{monotonicity} in context, local lexical-level entailment relations between a premise and hypothesis can be aggregated to determine the entailment relations at the sentence-pair level.  
For completeness of this paper, we highlight the key building blocks in Appendix~\ref{sec:background}. 

\subsection{\texttt{NatLogAttack} Setups and Principles}
Formally, given a premise sentence $P$, its $n$-word hypothesis $H=(h_{1},h_{2},\cdots, h_{n})$, and the ground-truth natural language inference label $y_g = \mathbb{L}(P, H)$,  \texttt{NatLogAttack} generates a hypothesis $H^{*}$ that satisfies a desired target label $y^{*}_g = \mathbb{L}(P, H^{*})$. 
The attacking pair $\langle P, H^{*} \rangle$ is generated only if the original pair $\langle P, H \rangle$ is correctly classified by a victim model $\mathbb{F}$. 
Accordingly, we denote $y = \mathbb{F}(P, H)$ as the natural language inference label predicated by the victim model $\mathbb{F}$ for the original pair and denote  $y^* = \mathbb{F}(P, H^*)$ as the predicted label for the attacking pair.

We propose to perform the attacks in two setups: the \textit{label-preserving} and  \textit{label-flipping} attacks. The attack principles and setups are summarized in Table~\ref{table:rules}. A \textit{label-preserving} attack generates adversarial examples with $y^{*}_g = y_g$, aiming to test the robustness of victim models on different inputs that have the same label---it attacks victim models under perturbations that do not change the inferential labels of the original premise-hypothesis pair. 

The \textit{label-flipping attacks}, on the other hand, aim at attacking victim models with perturbations that are key to differentiating two different logical relations where $y^{*}_g \neq y_g$. Note that natural logic can be naturally used to generate label-flipping attacks, and our work here is among the first to explore this type of attacks for natural language understanding, although label-flipping  attacks have been explored in image attacks ~\cite{tramer2020fundamental}. 






\begin{table}
\setlength{\tabcolsep}{1pt}
\centering
\resizebox{\linewidth}{!}{
\begin{tabular}{c c r@{\ }c@{\ }l@{} c}
\toprule
\textbf{Setups}    
&   \textbf{Label $\boldsymbol{y_g} \rightarrow \boldsymbol{y_g^{*}}$ }    
&   \multicolumn{3}{c}{\textbf{Strategy}}    
&   \textbf{Nat. Logic Relations}    
\\
\midrule
\multirow{3}{*}{Label-preserving} &   E $\rightarrow$ E  &    $H$&$\vDash$&$H^{*}$&   $H \equiv H^{*}$ or $H \sqsubset H^{*}$  \\
    &   C $\rightarrow$ C &  $H^{*} $&$\vDash$&$ H$ &   $H \equiv H^{*}$ or $H \sqsupset H^{*}$ \\
    &   N $\rightarrow$ N &   $H^{*} $&$\vDash$&$ H$
    &   $H \equiv H^{*}$ or $H \sqsupset H^{*}$\\
\midrule
\multirow{3}{*}{Label-flipping} &   E $\rightarrow$ C &  $H$&$ \vDash $&$\neg H^{*}$ &   $H \ \textsuperscript{$\wedge$} \ H^{*}$ or $H \mid H^{*}$  \\
    &   E $\rightarrow$ N &
    $H \nvDash H^{*}$ & and & $H \nvDash \neg H^{*}$
    &   $H \sqsupset\ H^{*}$ or $H \smallsmile H^{*}$
    \\ 
    &   C $\rightarrow$ E &  $\neg H^{*} $&$\vDash$&$ H$ &   $H \equiv \neg H^{*}$ or $H \sqsupset \neg H^{*}$ \\
\bottomrule
\end{tabular}
}
\caption{Generation principles of \texttt{NatLogAttack} and natural logic relations between the original hypothesis $H$ and the generated hypothesis $H^{*}$, where 
E, C and N stand for \textit{entailment}, \textit{contradiction} and \textit{neutral}.
}    
\label{table:rules}
\vspace{-1.5em}
\end{table}

The third column of the table (\textit{strategy}) lists the logic conditions between the generated hypothesis $H^*$ and the original hypothesis $H$ that satisfy the desired properties of preserving or flipping labels to obtain the target label $y_g^*$. Consider the second row of the label-preserving setup (\ie $C \rightarrow C$), in which \texttt{NatLogAttack} generates a hypothesis $H^{*}$ with $y_g^{*}=y_g=$~\textit{contradiction}. This is achieved by ensuring the natural language inference label between $H^{*}$ and $H$ to obey \textit{entailment}: $H^{*} \vDash H$.~\footnote{We use the \textit{entailment} notation that is same as in~\cite{maccartney2009extend}.} This guarantees the sentence pair $\langle P,H^{*} \rangle$ to have a \textit{contradiction} relation. In the natural logic formalism~\cite{maccartney2009natural}, this is implemented with $H \equiv H^{*}$ or $H \sqsupset H^{*}$. Consider another example. In the last row of the \textit{label-flipping} setup, \texttt{NatLogAttack} generates a new hypothesis $H^{*}$ with $y_g^{*}$~=~\textit{entailment} from a \textit{contradiction} pair, implemented by following the natural logic relations $H \equiv \neg H^{*}$ or $H \sqsupset \neg H^{*}$.

\newtheorem{assumption}{Constraint}[section]
\begin{assumption}
We constrain \texttt{NatLogAttack} from generating \textit{neutral} attack examples (\mbox{$y^*_g$= \textit{neutral}}) using the premise-hypothesis pairs with $y_g$=\textit{contradiction}, because two contradictory sentences may refer to irrelevant events from which a neutral pair cannot be reliably generated.~\footnote{For example, The SNLI~\cite{snli} and MNLI datasets ~\cite{mnli} were annotated under a guideline with a specific assumption of treating potentially irrelevant events as \textit{contraction}.}
\end{assumption}

\begin{assumption} 
\texttt{NatLogAttack} is also constrained from generating contradiction and entailment attacks (\mbox{$y^*_g$= \textit{contradiction}} or \mbox{$y^*_g$= \textit{entailment}}) from neutral pairs (\mbox{$y_g$=\textit{neutral}}), as there are many ways two sentences being \textit{neutral}, including reverse entailment and diverse semantic relations. 
The \textit{contradiction} and {entailment} pairs
cannot be reliably generated.
\end{assumption}

\subsection{Generation and Quality Control}
\label{sec:generation}


\subsubsection{Preparing Natural Logic Relations} 
As shown in the bottom-left part of Figure~\ref{fig:overview}, given a premise-hypothesis pair $\langle P, H \rangle$, the ground-truth label $y_g$, and the target label $y^*_g$, \texttt{NatLogAttack} retrieves natural logic relations from the last column of Table~\ref{table:rules}. Consider \textit{label-preserving} attacks and take $y_g^*=y_g=$\textit{entailment} as an example. From the last column in the first row of the \textit{label-preserving} setup, \texttt{NatLogAttack} finds and pushes the relations $\equiv$ and $\sqsubset$ into the \textit{natural-logic relations set},  $\frak{R}^*_g = \{\equiv,\sqsubset\}$, where $\frak{R}^*_g$ includes the natural-logic relations between $H$ and $H^*$ and will be used to generate the latter. Note that $r^*_g \in \frak{R}^*_g$ is one of relations in $\frak{R}^*_g$. 



We first copy $H$ to $H^{(1)}$, denoted as $H^{(1)}~\leftarrow~H$ for the convenience of notation, because the generation-and-attack process may be performed multiple rounds if one round of attacks fail. Then we use the notation $H^{(1)}$ and $H^{(2)}$ to refer to the original and a generated hypothesis sentence in each round. Note that in the above example, as will be discussed below, within each round of generation, \texttt{NatLogAttack} will provide a set of attacks to perform  multiple (iterative) attacks. 


\subsubsection{Candidate Generation} 

\newcommand\mycommfont[1]{\small\textcolor{blue}{#1}}
\SetCommentSty{mycommfont}
\SetKwInput{KwInput}{Input}             
\SetKwInput{KwOutput}{Output}      
\SetKw{KwInit}{Init}
\SetKwInput{KwReturn}{Return}  
\SetKwInput{KwParam}{Param}  
\begin{algorithm}[th!]
\SetAlgoLined
\DontPrintSemicolon
\small

\KwInput{Sentence~$H^{(1)}$ with tokens $(h^{(1)}_1,\cdots,h^{(1)}_n)$, target natural-logic relation set~$\frak{R}^*_g$}
\KwOutput{Candidate sentence set $\mathcal{H}$} 
\KwInit{$\mathcal{H} = \varnothing$}\\
\vspace{3pt}
$\frak{L} = \natlog(H^{(1)})$ \\
\vspace{3pt}
\ForEach{$h^{(1)}_i \in H^{(1)}$ and $r^*_g \in \frak{R}^*_g$}
{
  $\frak{R}^*_{local} = \frak{L}_\frak{B}[idx^{\frak{L}_i}(r^*_g)]$ \\
  
  \If{$\equiv \ \in \frak{R}^*_{local}$}
  {
    $\mathcal{H} = \mathcal{H} \ \cup $ PerturbSyno$(H^{(1)}, h_i^{(1)})$ \\
        $\mathcal{H} = \mathcal{H} \ \cup $ DoubleNegation$(H^{(1)})$ \\
  }
  
  \If{$\sqsubset \ \in \frak{R}^*_{local}$}
  {
        $\mathcal{H} = \mathcal{H} \ \cup $ PerturbHyper$(H^{(1)}, h_i^{(1)})$  \\
        $\mathcal{H} = \mathcal{H} \ \cup  $ Deletion$(H^{(1)},i)$  \\
   }
   
   \If{$\sqsupset \ \in \frak{R}^*_{local}$}
   {
        $\mathcal{H} = \mathcal{H} \ \cup $ PerturbHypo$(H^{(1)}, h_i^{(1)})$ \\
        $\mathcal{H} = \mathcal{H} \ \cup $ Insertion$(H^{(1)},i)$ 
    }
   \If{$\mid \ \in \frak{R}^*_{local}$}
    {
        $\mathcal{H} = \mathcal{H} \ \cup $ PerturbCoHyper$(H^{(1)}, h_i^{(1)})$ \\
        $\mathcal{H} = \mathcal{H} \ \cup $ PerturbAnto$(H^{(1)}, h_i^{(1)})$ \\
        $\mathcal{H} = \mathcal{H} \ \cup $ AltLM$(H^{(1)}, i)$ \\
    }    
    \If{$\wedge \ \in \frak{R}^*_{local}$}
    {
        $\mathcal{H} = \mathcal{H} \ \cup $  AddNeg$(H^{(1)}, h_i^{(1)})$ \\
    }
}
\KwReturn{$\mathcal{H}$}
\caption{Candidate Generation}\label{alg:gen_candidate}
\end{algorithm}

Our candidate attack generation process is described in
Algorithm~\ref{alg:gen_candidate}. Taking $H^{(1)}$ and $\frak{R}^*_g$ as the input, the algorithm aims to generate a set of candidate hypotheses $\mathcal{H}=\{H^{(2)}_1,\cdots,H^{(2)}_m\}$
with each pair $\langle H^{(1)}, H^{(2)}_i \rangle$ following a target relation $r^*_g \in \frak{R}^*_g$ where $H^{(2)}_i \in \mathcal{H}$. 
For each token $h^{(1)}_i \in H^{(1)}$ and $r^*_g \in \frak{R}^*_g$, 
the algorithm obtains the monotonicity and relation projection information using the Stanford \textit{natlog} parser\footnote{https://stanfordnlp.github.io/CoreNLP/natlog.html.} (\textit{line 2}). 
Specifically for $h_{i}^{(1)}$, suppose the parser outputs an ordered relation list: $\frak{L}_i = \langle \equiv, \sqsupset, \sqsubset, \wedge, \,\mid\,, \allowbreak \smallsmile, \# \rangle$, this returned list actually encodes the contextualized projection information, which we leverage to substitute $h_{i}^{(1)}$ with $h_{i}'$ to generate $H^{(2)}_i$ that satisfies relation $r^*_g$. 

In natural logic, when determining the sentence-level logical relation between a premise and hypothesis sentence, \textit{projection} is used to map local lexicon-level logical relation to sentence-level relations by considering the context and monotonicity. However, in adversarial attacks, \texttt{NatLogAttack} needs to take the following reverse action:  
\begin{align}
    \frak{R}_{local} &= 
    \frak{L}_\frak{B}
    [idx^{\frak{L}_i}(\boldmath{r^*_g})]
    \label{eq:rlocal}
\end{align}
where $r^*_g$ is the target sentence-level natural logic relation (in our above example, suppose $r^*_g$=`$\sqsubset$'). Then $idx^{\frak{L}_i}(.)$ returns the index of that relation in $\frak{L}_i$. For `$\sqsubset$', the index is 3. Then the index is used to find the lexicon-level (local) relation from  the predefined ordered list $\frak{L}_\frak{B}=\langle \,\equiv, \sqsubset, \sqsupset, \wedge, \,\mid\,,  \smallsmile,\allowbreak \#\,\rangle$. In the above example we will get $\frak{L}_\frak{B}[3]$=`$\sqsupset$'. Again, Equation~\ref{eq:rlocal} presents a reverse process of the regular \textit{projection} process in natural logic.
In other words, the ordered relation list provided by the \textit{natlog} parser for each word token, when used together with the predefined (ordered) relation list $\frak{L}_\frak{B}$, specifies a mapping between global (sentence-level) natural-logic relations and local (lexicon-level) relations. 
Note also that the output $\frak{R}_{local}$ is a set, because $\frak{L}_i$ is an ordered list that may contain the same relation multiple times.

\paragraph{Basic Word Perturbation.}
For a word token $h_i$, we replace it with word $h_i'$ to ensure the local relation $\langle h_i, h_i' \rangle$ to be $r_{local} \in \frak{R}_{local}$. 
\texttt{NatLogAttack} extracts natural-logic relation knowledge from knowledge bases to obtain word candidates for the desired relation types. 
The word perturbation of \texttt{NatLogAttack} focused on five relations in Table~\ref{table:basic-rel}.

\begin{assumption}
Since \textit{cover}~($\smallsmile$) is very rare and \textit{independence}~($\#$) is ambiguous, \texttt{NatLogAttack} is constrained to only focus on utilizing the remaining five relations: $\{\,\equiv, \sqsubset, \sqsupset, \wedge, \,\mid \}$. 
\end{assumption}

We attack the victim models using the most basic semantic relations explicitly expressed in knowledge bases and knowledge implicitly embedded in large pretrained language models. Specifically, we use WordNet~\cite{miller1995wordnet} to extract the desired lexical relations. For a word token $h_i$, we search candidate words $h_i'$ that has one of the following relations with \mbox{$h_i$: $\{\,\equiv, \\ \sqsubset, \sqsupset, \wedge, \,\mid \}$}. Synonyms are used as $h_i'$ to substitute $h_i$ for constructing $H^{(2)}$ with an \textit{equivalence} relation to $H^{(1)}$  (\textit{line 6}), hypernyms are used for \textit{forward entailment} \textit{(line 10)}, and hyponyms for \textit{reverse entailment} \textit{(line 14)}. 
Due to the transitiveness of \textit{forward entailment}~($\sqsubset$) and \textit{reverse entailment}~($\sqsupset$), we centre around $h_i$ to find its hypernyms and hyponyms but restrict the distances within a threshold to avoid generating sentences that are
semantically unnatural, contain overgeneralized concepts, or are semantically implausible. Later, we will further use a language model to control the quality. 

For \textit{alternation}, the perturbation candidates $h_i'$ are words that share the common hypernym with $h_i$ (\textit{line 18}). Following~\citet{maccartney2009natural}, we do not use antonyms of content words for the \textit{negation}  relation but instead use them to construct  \textit{alternation} hypotheses (\textit{line 19}). For the \textit{negation} (\textit{line 23}), a list of negation words and phrases is used to construct new hypotheses.
Note that while our experiments show the~\texttt{NatLogAttack} has been very effective and outperforms other attack  models, some of the components can be further augmented as future work.

\paragraph{Enhancing Alternation.} As discussed above, attacks may run multi-rounds if the prior round fails. For \textit{alternation} substitution, \texttt{NatLogAttack} does not replace the word token that has been substituted before, since the \textit{alternation} of \textit{alternation} does not guarantee to be the \textit{alternation} relation. In addition to constructing \textit{alternation} hypotheses using WordNet, we further leverage DistilBert~\cite{sanh2019distilbert} to obtain the alternation candidates using the function \textit{AltLM} (\textit{line 20}). Specifically, we mask the target word (which is a verb, noun, adjective or adverb) and prompt the language model to provide candidates. The provided candidates and replaced words are required to have the same POS tags. 

\begin{table}
\renewcommand\arraystretch{0.5}
\centering
\renewcommand\arraystretch{0.9}
\small
\begin{tabular}{ccc}
\toprule
\textbf{Monotonicity}    & \textbf{Upward}  & \textbf{Downward} \\

\midrule
\multirow{3}{*}{\textbf{Syntax}} &   \textit{adj}\ +\ \textit{n} $\sqsubset$ \textit{n}     &    \textit{adj}\ +\ \textit{n} $\sqsupset$ \textit{n} \\ 
    &   \textit{v}\ +\ \textit{adv} $\sqsubset$ \textit{v}  & \textit{v}\ +\ \textit{adv} $\sqsupset$ \textit{v}\\ 
    & \textit{s}\ +\ \textit{PP} $\sqsubset$ \textit{s} & \textit{s}\ +\ \textit{PP} $\sqsupset$ \textit{s}\\
\bottomrule
\end{tabular}
\caption{Insertion and deletion operations applied in the upward and downward context. \textit{s} is short for \textit{sentence}. }    
\label{tab:add_rules}
\vspace{-1.5em}
\end{table}




\paragraph{Insertion and Deletion.}
In addition to substitution, \texttt{NatLogAttack} also follows natural logic and monotonicity to construct examples using the insertion and deletion operations. As shown in Table~\ref{tab:add_rules}, adjectives, adverbs and prepositional phrases are leveraged in the upward and downward context of monotonicity to enhance the attacks for entailment (`$\sqsubset$') and reverse entailment (`$\sqsupset$'). 
We include the details in Appendix~\ref{sec:insertion}, which is built on Stanford \textit{CoreNLP} parser
and pretrained language models. Note that the syntactic rules do not guarantee to generate sentences with the desired NLI labels (e.g., see \cite{partee1995} for the discussion on the semantic composition of \textit{adjective}\ +\ \textit{noun}) and the process is only for generating candidates. We will use the pretrained language model to further identify good adversarial examples at a later stage.
Both the insertion and deletion operations are used with monotonicity and projection context to generate different relations.

\subsubsection{Attack Quality Control}

\texttt{NatLogAttack} uses DistilBert~\cite{sanh2019distilbert} to calculate the pseudo-perplexity scores~\cite{salazar2020masked} for all generated hypotheses $\mathcal{H} = \{H^{(2)}_{1},H^{(2)}_{2},\cdots, H^{(2)}_{m}\}$, 
and keeps only a maximum of 100 candidates with the lowest perplexity values. In our development, we found that the quality control stage is important for ensuring the quality of attack examples, particularly for reducing word perturbation mistakes resulting from incorrect interpretation of the words being substituted, which often results in unnatural hypothesis sentences, as well as reducing other sources of low-quality attacks including over-generalization of concepts and implausible semantics caused by insertion and deletion. The output of this stage is an ordered list of candidate attacks   
$\mathcal{H}_{sqc} = \langle H^{(2)}_{r_1},H^{(2)}_{r_2},\cdots, H^{(2)}_{r_k} \rangle$. 


\subsection{Iterative and Multi-rounds Attacking}



As discussed above, \texttt{NatLogAttack} performs iterative attacking within each round of generation and then multi-round attacks if the current round fails. Within each round, the original premise $P$ and each hypothesis in the ranked hypotheses list
$\mathcal{H}_{sqc}$ form an attack list $\langle \langle P,H^{(2)}_{r_1}\rangle,\cdots, \langle P,H^{(2)}_{r_k}\rangle \rangle$. As shown in Figure~\ref{fig:overview}, when an attack succeeds, we output the corresponding hypothesis as $H^*$, which is sent for evaluation. If an attack fails, the next pair in the ranked attack list will be tried until the list is exhausted. Then \texttt{NatLogAttack} organizes the next round of attacks. In total \texttt{NatLogAttack} generates a maximum of  500 attacks for each $\langle P,H \rangle$ pair.

When generating the next round attacks, we identify the adversarial  pair for which the victim model has the lowest confidence (indexed as $j_{\scriptscriptstyle lc}$) over the ground-truth class $y_{g}^{*}$:
\begin{align}
     j_{\scriptscriptstyle lc} &= \argmin\limits_{j \in \{r_1, \ldots, r_k\} } \{ s_{r_1}, \ldots, s_{r_k} \}   \\  
     s_{r_j} &= o(y_{g}^{*}|(P,H_{r_j}^{(2)}))
\vspace{-0.5em}
\end{align}
where $o(*)$ returns the corresponding softmax probabilities of the output layer. 
We then copy $H^{(2)}_{j_{\scriptscriptstyle lc}}$ to $H^{(1)}$, denoted as $ H^{(1)} \leftarrow H^{(2)}_{j_{\scriptscriptstyle lc}} $. The attack continues until the victim model is deceived to make a wrong prediction $y^{*}$ that is different from the ground truth $y_{g}^{*}$ or the maximum number of attacks is reached.

\section{Experiments and Results}
\subsection{Experimental Setup}
\label{sec:expsetup}
\paragraph{Dataset}
Our study uses SNLI~\cite{snli}, MNLI~\cite{mnli}, MED~\cite{med2019}, HELP~\cite{help2019}, and SICK~\cite{sick,hu2020monalog} datasets. The MED upward and downward subsets are denoted as $\text{MED}_{\text{up}}$ and  $\text{MED}_{\text{down}}$, respectively.  Details of the datasets and the setup for training can be found in Appendix~\ref{sec:baselineDetails}. 



\begin{table}
\renewcommand\arraystretch{0.6}
\small
    \centering
    \setlength{\tabcolsep}{3pt}
    \begin{tabular}{ccccccc}
        \toprule
        Models & SNLI & MED & $\text{MED}_{\text{up}}$ & $\text{MED}_{\text{down}}$ & MNLI & SICK \\
        \midrule
        BERT  & 89.99 & 77.68 & 74.42 & 81.72 & 84.32 & 87.06 \\ 
        RoBERTa  & 91.53 & 73.37 & 80.97 & 70.72 & 87.11 & 87.79\\ 
        \bottomrule
    \end{tabular}
    \caption{Victim models' accuracy on different datasets.}
\label{tab:dataset}
 \vspace{-1.5em}
\end{table}

\paragraph{Attack and Victim Models}
We compared the proposed model to five representative attack models including the recent state-of-the-art models: \texttt{Clare}~\cite{li2021clare}, \texttt{BertAttack}~\cite{li2020bertattack}, \texttt{PWWS}~\cite{pwws}, \texttt{TextFooler}~\cite{textfooler} and \texttt{PSO}~\cite{pso}.
Specifically, we used the implementation made publicly available in \texttt{TextAttack}.\footnote{https://github.com/QData/TextAttack} For victim models, we used uncased \texttt{BERT}~\cite{bert} and \texttt{RoBERTa} base models~\cite{roberta}. The accuracy of victim models is included in Table~\ref{tab:dataset}, which is comparable to the state-of-the-art performance. 



    




\paragraph{Evaluation Metrics} Three metrics are used to evaluate the models from different perspectives. The sign $\uparrow$ ($\downarrow$) indicates that the higher (lower) the values are, the better the performance is.

\begin{table*}[th]
\renewcommand\arraystretch{1.1}
    \centering
    \setlength{\tabcolsep}{3pt}
    \resizebox{1\linewidth}{!}{
    \begin{tabular}{c|l|rrr|rrr|rrr|rrr|rrr|rrr}
    \toprule
        \multirow{2}{*}{Victim} & \multicolumn{1}{c|}{\multirow{2}{*}{Attack}} & \multicolumn{3}{c|}{SNLI}& \multicolumn{3}{c|}{MED} & \multicolumn{3}{c|}{$\text{MED}_{\text{up}}$}&\multicolumn{3}{c|}{$\text{MED}_{\text{down}}$}& \multicolumn{3}{c|}{$\text{MNLI}$}&\multicolumn{3}{c}{$\text{SICK}$}\\
        \cmidrule{3-20}
        Model&\multicolumn{1}{c|}{Model}&\small{HVASR} &  QN & PPL  &\small{HVASR} &  QN & PPL &\small{HVASR} & QN & PPL &\small{HVASR} &  QN & PPL&\small{HVASR} & QN & PPL &\small{HVASR} &  QN & PPL\\
        \midrule
      \multirow{6}{*}{\makecell[c]{\rotatebox[origin=c]{90}{BERT}}}  
      & PWWS             & 29.9   & 175.8 & 15.96     & 45.9 & 115.3    & 18.18        &  43.1    &  119.1    &  17.98       &  \underline{48.3}    &  111.6    &  18.38 & 27.8 & 184.2 & 13.87 &  31.0  & 147.1  & 17.75\\
      & Textfooler       &  \underline{34.5}  & \underline{58.4}  &  \underline{15.88}    & \underline{47.3} & \underline{51.2}     & \underline{17.96}     &  \underline{47.8}    &  \underline{51.2}    &  \underline{17.77}       &  46.9    &  \textbf{51.2}    &  \underline{18.15}  & 37.3 & \underline{74.7} & \underline{13.62} &   30.7  &  \underline{50.0}  & \underline{17.62}\\
      & PSO             &  20.5  & 91.8  & 16.06      & 38.8 & 81.9     & 18.19      &  37.7    &  83.9    &  18.14     &  39.7    &  79.7    &  18.25   & 32.0 & 103.4 & 13.81 &  22.3    &   115.86  & 17.77\\
      & BertAttack      &  31.6  &  76.4 &  17.07    & 39.9 & 62.3     & 18.86    &   31.1   &  63.2    &  18.7      &    47.4  &  61.5    &  19.02   &  \underline{37.4}  & 86.5  & 14.47  &  \underline{32.2}   & 91.7  & 18.18\\
      & Clare           &   19.9 & 328.3 &  16.87    & 36.7 & 199.7    & 18.31       &    29.9  &  205.5    &  18.30       &  42.8    &  194.8    &  18.33 &  25.2  &  299.8  & 16.87   &    23.1     & 246.9  & 18.60\\
      & NatLogAtt*    &  \textbf{35.7}  &  \textbf{42.8} &  \textbf{14.78}    & \textbf{56.9}  & \textbf{42.7}     &  \textbf{17.43}       &  \textbf{57.9}    &  \textbf{30.1}    &  \textbf{17.24}       &  \textbf{56.0}    &  \underline{55.4}    &  \textbf{17.62}   &  \textbf{39.7}  & \textbf{50.1}  & \textbf{13.47}  &   \textbf{43.6}    & \textbf{40.3}  & \textbf{16.73}\\
      \midrule
      \multirow{6}{*}{\makecell[c]{\rotatebox[origin=c]{90}{RoBERTa}}}
 
      & PWWS     & \underline{35.5} & 177.1 & 16.05 & 39.8 & 118.5 & 18.15 & 41.3 & 121.1 & 18.30 & 38.7 & 115.8 & 18.00 & 28.7  & 189.6  &  13.83  & 35.2   &  143.4  & 17.91\\ 
      & Textfooler & 30.0 & \underline{59.7} & \underline{15.93}  & 42.6 & \underline{50.2} & \underline{18.06}& 38.7 & \underline{49.5} & \underline{17.98}  & 45.6 & \underline{50.82} & \underline{18.13} &  34.0 & \underline{78.2}  & \underline{13.61}  & 33.8     & \underline{49.6}  & \underline{17.69} \\
      & PSO  & 19.2 & 92.9 & 16.17 & 34.3 & 81.8 & 18.14  & 27.1 & 83.2 & 18.03  & 39.3 & 80.19 & 18.26 & 28.3  & 99.4  & 13.85 &    24.9  &115.0  & 17.75\\ 
      & BertAttack  & 34.9 & 78.3 & 16.89  & \underline{47.3} & 61.1 & 18.77 & \underline{47.2} & 59.7 & 18.66  & \underline{47.4} & 62.4 & 18.89 & \underline{39.2}  &91.2  & 14.65  &   \underline{35.6}    &95.8  & 18.21\\ 
      & Clare  & 14.7 & 326.6 & 16.65  &27.4 & 199.8 & 18.54  & 17.9 & 203.7 & 18.20  & 35.2 & 195.9 & 18.88 & 22.6  & 296.7  &  16.44  &  27.5    & 244.0  & 18.16\\ 
      & NatLogAtt*  & \textbf{36.5} & \textbf{45.0} & \textbf{14.69} & \textbf{55.5} & \textbf{33.9} & \textbf{17.37}  & \textbf{59.7} & \textbf{27.5} & \textbf{17.34}  & \textbf{52.3} & \textbf{40.2} & \textbf{17.40} &  \textbf{39.7}  & \textbf{46.1}  & \textbf{13.53} &   \textbf{49.3}    & \textbf{42.9}  & \textbf{16.61}\\ 
         \bottomrule
    \end{tabular}
    }
    \caption{Performance of different attack models in label-preserving attacks. 
    The bold font marks the best performance under each evaluation setup. 
    The improvements of \texttt{NatLogAtt} over the second-best results (marked with underscores) are statistically significant ($p<0.05$) under one-tailed paired t-test. 
    }  
    \label{tab:res1}
    \vspace{-1em}
\end{table*}

\vspace{-0.5em} 
\begin{itemize}[leftmargin=3.0mm]
\setlength\itemsep{-1.0mm}
\item \textbf{Human Validated Attack Success Rate (HVASR $\uparrow$).} 
Most existing attacking methods are evaluated with attack success rates that are not validated by human subjects, assuming that the attacking methods could generate adversarial examples of the desired labels. This assumption works for many NLP tasks such as sentiment analysis and text classification.  However, this is not the case in NLI, since the logical relationships can be easily broken during the generation process. As observed in our experiments, although the state-of-art attacking models (\texttt{BertAttack} and \texttt{Clare}) attain high attack success rates on various NLP tasks, human-validated evaluation demonstrates that they are much less effective in attacking natural language reasoning.  
To reliably evaluate the attack performance, we use \textit{Human Validated Attack Success Rate} (HVASR). 
Specifically, we used Amazon Mechanical Turk\footnote{https://www.mturk.com/} to validate if the generated attack examples belong to the desired relations. Each example was annotated by at least three workers and the label is determined by the majority voting.
HVASR is the percentage of \textit{successful-and-valid} adversarial examples that successfully deceived the victim models to make the wrong prediction and at the same time the majority of the annotators think their NLI labels are the desired target labels $y^{*}_g$. 
 While HVASR is our major evaluation metric, we also use query numbers and perplexity to provide additional perspectives for observations.

    \item \textbf{Query number~(QN $\downarrow$)} refers to the average number of times that a successful attack needs to query the victim model~\cite{pso,li2020bertattack}. QN can reflect the efficiency (but not effectiveness) of an attack model.
    
    \item \textbf{Perplexity~(PPL $\downarrow$)} reflects the fluency and quality of generated examples. Same as in~\cite{pso,li2021clare}, it is  computed with GPT-2~\cite{gpt2} during evaluation. 
\end{itemize}

\subsection{Results and Analysis}


\paragraph{Results on Label Preserving Attacks}
Table~\ref{tab:res1} shows the performance of different models on \textit{label-preserving attacks}. We can see that \texttt{NatLogAttack} consistently achieves the best performance on HVASR. 
The detailed results on MED also show that \texttt{NatLogAttack} has a better ability to construct adversarial examples in both upward and downward monotone. \texttt{NatLogAttack} also shows superior performance on average QN and PPL in nearly all setups. 

We can see that \texttt{NatLogAttack} has a large HVASR and small QN value in $\text{MED}_{\text{up}}$, suggesting that \texttt{NatLogAttack} can easily generate attacks in the upward monotone. However, in $\text{MED}_{\text{down}}$, \texttt{NatLogAttack} needs more efforts (QN). Our further analysis reveals that this is because in the downward monotone, the attack model relies more on the insertion operation than deletion, and the former is more likely to result in unsuccessful attempts. 

\begin{figure}[t]
\centering

\begin{minipage} [t]{1\linewidth}
\centering
    \includegraphics[width=1\linewidth,clip]{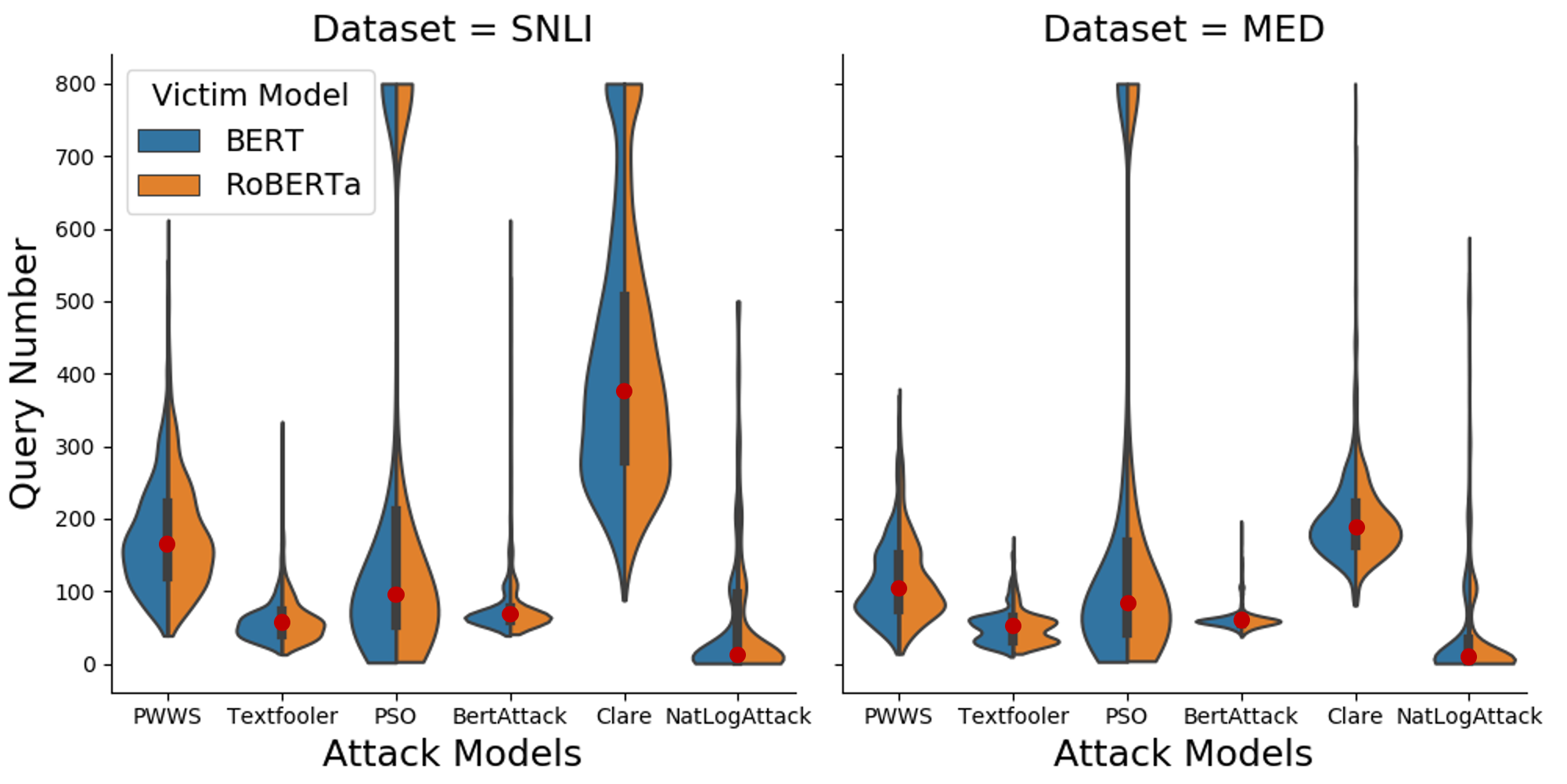}
\end{minipage}

\caption{Query numbers (QNs) of attack models. Red dots are the medians of QNs of different attack models. The blue and orange shapes show the densities of query numbers for \texttt{BERT} and \texttt{RoBERTa}, respectively.
}

\label{fig:q_num}
\end{figure}

Figure~\ref{fig:q_num} further compares the query numbers (QNs) of different attack models on \texttt{BERT} and \texttt{RoBERTa} in terms of the medians (instead of means) and density of QN.
We can see that the majority of query numbers of \texttt{NatLogAttack} are rather small and medians are less than 12 for on both SNLI and MED, showing that \texttt{NatLogAttack} could attack successfully with very limited attempts in most cases.
For each attack model, the density of QN on \texttt{BERT} and \texttt{RoBERTa}
is close to each other and the medians are indiscernible and are represented by the same red dot in the figure.

\begin{table}
\renewcommand\arraystretch{1.2}
    \centering
    \setlength{\tabcolsep}{1pt}
    \resizebox{1\linewidth}{!}{
    \begin{tabular}{c|c|ccc|ccc|ccc|ccc}
    \toprule
        \multirow{2}{*}{Vict.} & \multirow{2}{*}{Lab.} & \multicolumn{3}{c|}{SNLI}&\multicolumn{3}{c|}{MED}&\multicolumn{3}{c|}{\text{MNLI}}&\multicolumn{3}{c}{SICK}\\
        \cmidrule{3-14}
        Md. &Flip. &\scriptsize{HVASR} & \small{QN}  & \small{PPL} & \scriptsize{HVASR}  & \small{QN}  & \small{PPL}& \scriptsize{HVASR}  & \small{QN}  & \small{PPL}& \scriptsize{HVASR}  & \small{QN}  & \small{PPL} \\
        \midrule
      \multirow{3}{*}{\makecell[c]{\rotatebox[origin=c]{90}{BERT}}} 

       & E$\shortrightarrow$C  & 37.9 & 1.0{ } & 14.8  & 48.7 & 1.0{ } & 16.9 & 33.2  & 1.4{ } & 13.5 & 31.8  & 10.4{ } & 16.2\\ 
       & E$\shortrightarrow$N & 57.5 & 2.9{ } & 14.9  & 50.9 & 2.8{ } & 17.7 &  50.3 & 4.7{ } & 13.7 & 55.8  & 6.5{ } & 16.1\\ 
       & C$\shortrightarrow$E  & 33.4 & 1.0{ } &  14.4 & - & - & - & 34.2 & 1.1{ } & 13.0 & 37.1  & 1.0{ } & 16.0\\                
      \midrule
      \multirow{3}{*}{\makecell[c]{\rotatebox[origin=c]{90}{RoBERTa}}}  

      & E$\shortrightarrow$C  & 43.5 & 1.4{ } & 14.6 & 49.8 & 2.9{ } & 16.7 & 36.8  & 5.0{ } & 13.5 & 32.1  & 13.9{ } & 16.4\\ 
      & E$\shortrightarrow$N  & 56.8 & 2.6{ } & 14.8 & 52.1 & 3.0{ } &  17.6 & 50.7 & 4.8{ } & 13.8 & 57.4  & 4.4{ } & 16.1\\ 
      & C$\shortrightarrow$E  & 36.4 & 1.8{ } & 14.5 & - & - & - & 35.1 & 1.2{ } & 13.0 & 37.7  & 1.0{ } & 16.0\\
      \bottomrule
    \end{tabular}
    }
    \caption{The evaluation for label-flipping attacks. 
    }
    \label{tab:res_flip}
    \vspace{-1.5em}
\end{table}

\begin{table*}[!t]
\renewcommand\arraystretch{0.75}
\centering
\resizebox{1\linewidth}{!}{
    \centering
    \begin{tabular}{c|c|c}
         \toprule
         Attack Model   &  Premise &  Hypothesis  \\
        
        
        
         
         \midrule  
         
         PWWS  & Betty lives in Berlin &         Betty \st{lives} \textcolor{blue}{animation} in Germany  \\ 
         Textfooler &     Betty lives in Berlin &         Betty \st{lives} \textcolor{blue}{dies} in Germany  \\
         PSO &     - &         -  \\
         BertAttack  & Betty lives in \st{Berlin} \textcolor{blue}{prague}   & Betty lives in Germany  \\ 
         Clare  & Betty lives in \st{Berlin} \textcolor{blue}{Australia}   & Betty lives in Germany  \\ 
         NatLogAttack  & Betty lives in Berlin &         Betty lives in \st{Germany} \textcolor{blue}{Federal Republic of Germany} \\
         
         \midrule  
         PWWS &    A snow \st{goose} \textcolor{blue}{jackass} is a water bird &    A goose is a water bird  \\
         Textfooler &    A snow goose is a water bird &         A goose is a water \st{bird} \textcolor{blue}{parakeets}  \\
         PSO &    A snow goose is a water bird &         A \st{goose} \textcolor{blue}{chicken} is a water bird  \\
         BertAttack & A snow \st{goose} \textcolor{blue}{the} is a water bird 	& A goose is a water bird \\
         Clare &    A snow \st{goose} \textcolor{blue}{cat} is a water bird &      A goose is a water bird  \\
         NatLogAttack &    A snow goose is a water bird &         A goose is a \st{water} \underline{\st{bird}} \textcolor{blue}{chordate}  \\ 
         
        
        \midrule 
        
        PWWS  & - &         -  \\ 
         Textfooler &     - &         -  \\
         PSO &     - &         -  \\
         BertAttack  & I can't speak German at all   & I \st{can't} \textcolor{blue}{cantheisland} speak \st{German confidently} \textcolor{blue}{and never} at all  \\ 
         Clare  & I can't speak German at all  & I \st{can't speak} \textcolor{blue}{spoke} German confidently at all  \\ 
         NatLogAttack  & I can't speak German at all &        I  can't speak German confidently at all \textcolor{red}{on trampoline} \\
         
         \midrule 
        
        PWWS  & The \st{purple} \textcolor{blue}{majestic} alien did not throw balls
 &         The purple alien did not throw tennis balls  \\ 
         Textfooler &     The purple alien did not throw balls
 &         The \st{purple} \textcolor{blue}{crimson} alien did not throw \st{tennis} \textcolor{blue}{opening} balls  \\
         PSO &     The purple alien did not throw balls
 &         The purple \st{alien} \textcolor{blue}{unicorn} did not throw tennis balls  \\
         BertAttack  & The \st{purple} \textcolor{blue}{blue} alien did not throw balls
   & The purple alien did not throw tennis balls  \\ 
         Clare  & The purple alien did not throw \textcolor{red}{soccer} balls
  & The purple alien did not throw balls  \\ 
         NatLogAttack  & The purple alien did not throw balls
 &        The purple alien did not throw tennis balls  \textcolor{blue}{on her cellphone} \\
       
         \bottomrule
    \end{tabular}
    }
    \caption{Adversarial examples generated by different attack models on MED under the \textit{label-preserving} setup (E $\rightarrow$ E). The victim model is RoBERTa. Insertion is marked in red, substitution in blue, and deletion is marked with underline.
    The symbol `-' indicates that the attack model fails to generate examples. The top two groups of examples are upward monotone and the bottom two groups are downward monotone.}
    \label{tab:case1}
    \vspace{-1em}
\end{table*}

\begin{table*}[!t]
\setlength{\tabcolsep}{5pt}
\centering
\resizebox{1\linewidth}{!}{
    \centering
    \begin{tabular}{c|c|c}
         \toprule
         Label Flip.    &  Premise &  Hypothesis   \\
        
        \midrule 
         \multirow{2}{*}{E $\rightarrow$ C}& Many aliens drank some coke &	Many aliens drank some \st{soda} \textcolor{blue}{alcohol}   \\
         & He lied, without hesitation &	He \st{lied} \textcolor{blue}{did not lie}, without any hesitation  \\
         
        \midrule 
        \multirow{2}{*}{E $\rightarrow$ N} & She's wearing a nice big hat &	She's wearing a nice \textcolor{red}{straw} hat   \\
         & Two formally dressed, bald older women &	Two bald \st{women} \textcolor{blue}{matriarchs}   \\
        
        \midrule  
          \multirow{2}{*}{C $\rightarrow$ E} &    A little boy is riding a yellow bicycle across a town square &         \textcolor{red}{It is false that} the boy's bike is blue  \\
          &    Two men in orange uniforms stand before a train and do some work &     \textcolor{red}{It is not true that}  nobody is working  \\
         \bottomrule
         
    \end{tabular}
    }
    \caption{Adversarial examples generated by the \texttt{NatLogAttack} model in the \textit{label-flipping} setup. The victim model is RoBERTa. The red and blue colours highlight the insertion or substitution, respectively.}   

    \label{tab:case2}
\end{table*}

\paragraph{Results on Label Flipping Attacks}
 Table~\ref{tab:res_flip} shows the performance of \texttt{NatLogAttack} on the \textit{label-flipping attacks}. Note that there has been little prior work providing systematic label-flipping attacks for NLP tasks. This new angle of evaluation is more easily implemented with logic-based attacks and provides additional insights. 
 Specifically, the table shows that the numbers of queries that \texttt{NatLogAttack} sent to the victim models are much smaller than those in the \textit{label-preserving} setting presented in Table~\ref{tab:res1}, suggesting that the victim models are more vulnerable in \textit{label-flipping setting}. For example, we can see that most of the query numbers are within 1-5 in Table~\ref{tab:res_flip}. The pretrained victim models are capable of memorizing the superficial features related to the original label and have difficulty  in capturing the logical relationship when we alter them between sentences by keeping the majority of words untouched. 

 
 In both the \textit{label-preserving} and \textit{label-flipping} setup, the HVASR may still be further improved, although the proposed models have substantially outperformed the off-the-shelf state-of-the-art attack models and cautions have been exercised in all attack generation steps, which leaves  room for more research on improving logic-based attacks as future work. 

\paragraph{Examples and Analysis.}
Table~\ref{tab:case1} provides the generated attack examples in the \textit{label-preserving} setup ($E \rightarrow E$), in which we can see the quality of attacks generated by \texttt{NatLogAttack} is clearly higher.
The baseline attacking models generate adversarial examples by replacing words based on word embedding or language models, which can easily break the logic relationships.
Some examples in Table~\ref{tab:case1} show that the baselines often rely on semantic \textit{relatedness} to construct adversarial examples, which is not detailed enough for NLI and hence break the logic relations (e.g., the last \texttt{BertAttack} example). Also, the last example of \texttt{Clare} shows that the model deletes words without considering the context (downward) monotonicity, resulting in an invalid attack. Note that the baseline models modify both premises and hypotheses and \texttt{NatLagAttack} focuses only on modifying hypotheses---it is straightforward to copy or adapt the operations of \texttt{NatLagAttack} to modify premises---in many applications, it is more natural to modify the hypotheses and keep the premises (evidences) untouched.

Table~\ref{tab:case2} shows more adversarial examples generated by \texttt{NatLogAttack} in the \textit{label-flipping} setup. For all the six examples, the prediction of the victim model \texttt{RoBERTa} remains unchanged (\ie \textit{entailment}, \textit{entailment} and \textit{contradiction} for the first, middle, and last two examples, respectively), while the ground-truth labels are now \textit{contradiction}, \textit{neutral}, and \textit{entailment}, respectively. The victim model had difficulty in telling the difference, which renders an angle to challenge the models' ability of understanding and perform reasoning.



\section{Conclusion}
Towards developing logic-based attack models, we introduce a framework \texttt{NatLogAttack}, which centres around the classical natural logic formalism. The experiments with human and automatic evaluation show that the proposed framework outperforms the existing attack methods. Compared to these models, \texttt{NatLogAttack} generates better adversarial examples with fewer visits to the victim models. In addition to the widely used label-preserving attacks, \texttt{NatLogAttack} also provides label-flipping attacks. The victim models are found to be more vulnerable in this setup and \texttt{NatLogAttack}
succeeds in deceiving them with much smaller
numbers of queries. 
\texttt{NatLogAttack} provides an approach to probing the existing and future NLI models' capacity from a key viewpoint and we hope more logic-based attacks will be
further explored for understanding the desired property of reasoning. 

\section*{Limitations}
Our research focuses on the adversarial attack itself and provides a framework that can be potentially used in different adversarial training strategies. We limit ourselves on attacks in this work, but it would be interesting to investigate logic-based attacks in adversarial training. We will leave that as future work. The proposed attack approach is also limited by the limitations of natural logic, while the latter has been a classical logic mechanism. For example, our proposed framework has less deductive power than first-order logic. It cannot construct attacks building on inference rules like \textit{modus ponens}, \textit{ modus tollens}, and \textit{disjunction elimination}. As discussed in the paper, some components of the generation and quality control process can be further enhanced.

\section*{Acknowledgements}
The research is supported by the NSERC Discovery Grants and the Discovery Accelerator Supplements. We thank Bairu Hou for his contributions to an early version of the proposed model.

\bibliography{anthology,natlog}

\begin{thebibliography}{72}
\expandafter\ifx\csname natexlab\endcsname\relax\def\natexlab#1{#1}\fi

\bibitem[{Alzantot et~al.(2018)Alzantot, Sharma, Elgohary, Ho, Srivastava, and
  Chang}]{ga}
Moustafa Alzantot, Yash Sharma, Ahmed Elgohary, Bo-Jhang Ho, Mani Srivastava,
  and Kai-Wei Chang. 2018.
\newblock \href {https://doi.org/10.18653/v1/D18-1316} {Generating natural
  language adversarial examples}.
\newblock In \emph{Proceedings of the 2018 Conference on Empirical Methods in
  Natural Language Processing (EMNLP)}, pages 2890--2896, Brussels, Belgium.
  Association for Computational Linguistics.

\bibitem[{Angeli and Manning(2014)}]{angeli2014naturalli}
Gabor Angeli and Christopher~D Manning. 2014.
\newblock Naturalli: Natural logic inference for common sense reasoning.
\newblock In \emph{Proceedings of the 2014 conference on empirical methods in
  natural language processing (EMNLP)}, Doha, Qatar.

\bibitem[{Angeli et~al.(2016)Angeli, Nayak, and Manning}]{angeli2016combining}
Gabor Angeli, Neha Nayak, and Christopher~D Manning. 2016.
\newblock Combining natural logic and shallow reasoning for question answering.
\newblock In \emph{Proceedings of the 54th Annual Meeting of the Association
  for Computational Linguistics (ACL)}, Berlin, Germany.

\bibitem[{Belinkov et~al.(2019)Belinkov, Poliak, Shieber, Van~Durme, and
  Rush}]{belinkov2019don}
Yonatan Belinkov, Adam Poliak, Stuart~M Shieber, Benjamin Van~Durme, and
  Alexander~M Rush. 2019.
\newblock Don’t take the premise for granted: Mitigating artifacts in natural
  language inference.
\newblock In \emph{Proceedings of the 57th Annual Meeting of the Association
  for Computational Linguistics (ACL)}, pages 877--891.

\bibitem[{Bowman et~al.(2015)Bowman, Angeli, Potts, and Manning}]{snli}
Samuel~R. Bowman, Gabor Angeli, Christopher Potts, and Christopher~D. Manning.
  2015.
\newblock \href {https://doi.org/10.18653/v1/D15-1075} {A large annotated
  corpus for learning natural language inference}.
\newblock In \emph{Proceedings of the 2015 Conference on Empirical Methods in
  Natural Language Processing (EMNLP)}, pages 632--642, Lisbon, Portugal.
  Association for Computational Linguistics.

\bibitem[{Chen et~al.(2018)Chen, Zhu, Ling, Inkpen, and Wei}]{kim2018}
Qian Chen, Xiaodan Zhu, Zhen-Hua Ling, Diana Inkpen, and Si~Wei. 2018.
\newblock Neural natural language inference models enhanced with external
  knowledge.
\newblock In \emph{Proceedings of the 56th Annual Meeting of the Association
  for Computational Linguistics (ACL)}, Melbourne, Australia.

\bibitem[{Chen et~al.(2017)Chen, Zhu, Ling, Wei, Jiang, and
  Inkpen}]{Qian2017esim}
Qian Chen, Xiaodan Zhu, Zhenhua Ling, Si~Wei, Hui Jiang, and Diana Inkpen.
  2017.
\newblock Enhanced lstm for natural language inference.
\newblock In \emph{Proceedings of the 55th Annual Meeting of the Association
  for Computational Linguistics (ACL)}, Vancouver. ACL.

\bibitem[{Chen and Gao(2021)}]{chen2021monotonicity}
Zeming Chen and Qiyue Gao. 2021.
\newblock Monotonicity marking from universal dependency trees.
\newblock In \emph{Proceedings of the 14th International Conference on
  Computational Semantics (IWCS)}, pages 121--131.

\bibitem[{Chen et~al.(2021)Chen, Gao, and Moss}]{chen2021neurallog}
Zeming Chen, Qiyue Gao, and Lawrence~S Moss. 2021.
\newblock Neurallog: Natural language inference with joint neural and logical
  reasoning.
\newblock In \emph{Proceedings of* SEM 2021: The Tenth Joint Conference on
  Lexical and Computational Semantics}, pages 78--88.

\bibitem[{Dagan et~al.(2005)Dagan, Glickman, and
  Magnini}]{DBLP:conf/mlcw/DaganGM05}
Ido Dagan, Oren Glickman, and Bernardo Magnini. 2005.
\newblock The {PASCAL} recognising textual entailment challenge.
\newblock In \emph{Proceedings of the First international conference on Machine
  Learning Challenges: evaluating Predictive Uncertainty Visual Object
  Classification, and Recognizing Textual Entailment}.

\bibitem[{De~Raedt et~al.(2019)De~Raedt, Manhaeve, Dumancic, Demeester, and
  Kimmig}]{de2019neuro}
Luc De~Raedt, Robin Manhaeve, Sebastijan Dumancic, Thomas Demeester, and
  Angelika Kimmig. 2019.
\newblock Neuro-symbolic= neural+ logical+ probabilistic.
\newblock In \emph{NeSy'19@ IJCAI, the 14th International Workshop on
  Neural-Symbolic Learning and Reasoning}, Macao, China.

\bibitem[{Devlin et~al.(2019)Devlin, Chang, Lee, and Toutanova}]{bert}
Jacob Devlin, Ming-Wei Chang, Kenton Lee, and Kristina Toutanova. 2019.
\newblock \href {https://doi.org/10.18653/v1/N19-1423} {{BERT}: Pre-training of
  deep bidirectional transformers for language understanding}.
\newblock In \emph{Proceedings of the 2019 Conference of the North {A}merican
  Chapter of the Association for Computational Linguistics: Human Language
  Technologies, Volume 1 (Long and Short Papers)}, pages 4171--4186,
  Minneapolis, Minnesota. Association for Computational Linguistics.

\bibitem[{Dong et~al.(2010)Dong, Dong, and Hao}]{dong2006hownet}
Zhendong Dong, Qiang Dong, and Changling Hao. 2010.
\newblock \href {https://aclanthology.org/C10-3014} {{H}ow{N}et and its
  computation of meaning}.
\newblock In \emph{Coling 2010: Demonstrations}, pages 53--56, Beijing, China.
  Coling 2010 Organizing Committee.

\bibitem[{Ebrahimi et~al.(2018)Ebrahimi, Rao, Lowd, and
  Dou}]{ebrahimi2017hotflip}
Javid Ebrahimi, Anyi Rao, Daniel Lowd, and Dejing Dou. 2018.
\newblock \href {https://doi.org/10.18653/v1/P18-2006} {{H}ot{F}lip: White-box
  adversarial examples for text classification}.
\newblock In \emph{Proceedings of the 56th Annual Meeting of the Association
  for Computational Linguistics (Volume 2: Short Papers)}, pages 31--36,
  Melbourne, Australia. Association for Computational Linguistics.

\bibitem[{Evans and Grefenstette(2018)}]{evans2018learning}
Richard Evans and Edward Grefenstette. 2018.
\newblock Learning explanatory rules from noisy data.
\newblock In \emph{Journal of Artificial Intelligence Research (JAIR)},
  volume~61, pages 1--64.

\bibitem[{Feng et~al.(2022)Feng, Yang, Greenspan, and Zhu}]{Feng2022}
Yufei Feng, Xiaoyu Yang, Michael Greenspan, and Xiaodan Zhu. 2022.
\newblock Neuro-symbolic natural logic with introspective revision for natural
  language inference.
\newblock \emph{Transactions of the Association for Computational Linguistics
  (TACL)}, 10:240–256.

\bibitem[{Feng et~al.(2020)Feng, Zheng, Liu, Greenspan, and
  Zhu}]{feng2020exploring}
Yufei Feng, Zi'ou Zheng, Quan Liu, Michael Greenspan, and Xiaodan Zhu. 2020.
\newblock Exploring end-to-end differentiable natural logic modeling.
\newblock In \emph{Proceedings of the 28th International Conference on
  Computational Linguistics (COLING)}, pages 1172--1185.

\bibitem[{Garcez et~al.(2015)Garcez, Besold, De~Raedt, F{\"o}ldiak, Hitzler,
  Icard, K{\"u}hnberger, Lamb, Miikkulainen, and Silver}]{garcez2015neural}
Artur~d'Avila Garcez, Tarek~R Besold, Luc De~Raedt, Peter F{\"o}ldiak, Pascal
  Hitzler, Thomas Icard, Kai-Uwe K{\"u}hnberger, Luis~C Lamb, Risto
  Miikkulainen, and Daniel~L Silver. 2015.
\newblock Neural-symbolic learning and reasoning: contributions and challenges.
\newblock In \emph{2015 AAAI Spring Symposium Series}.

\bibitem[{Geiger et~al.(2020)Geiger, Richardson, and Potts}]{geiger2020neural}
Atticus Geiger, Kyle Richardson, and Christopher Potts. 2020.
\newblock Neural natural language inference models partially embed theories of
  lexical entailment and negation.
\newblock In \emph{Proceedings of the Third BlackboxNLP Workshop on Analyzing
  and Interpreting Neural Networks for NLP}, pages 163--173.

\bibitem[{Glockner et~al.(2018)Glockner, Shwartz, and Goldberg}]{breakingNLI}
Max Glockner, Vered Shwartz, and Yoav Goldberg. 2018.
\newblock \href {https://doi.org/10.18653/v1/P18-2103} {Breaking {NLI} systems
  with sentences that require simple lexical inferences}.
\newblock In \emph{Proceedings of the 56th Annual Meeting of the Association
  for Computational Linguistics (Volume 2: Short Papers)}, pages 650--655,
  Melbourne, Australia. Association for Computational Linguistics.

\bibitem[{Goodfellow et~al.(2014)Goodfellow, Shlens, and
  Szegedy}]{goodfellow2014explaining}
Ian~J Goodfellow, Jonathon Shlens, and Christian Szegedy. 2014.
\newblock Explaining and harnessing adversarial examples.
\newblock \emph{arXiv preprint arXiv:1412.6572}.

\bibitem[{Hu et~al.(2020)Hu, Chen, Richardson, Mukherjee, Moss, and
  Kuebler}]{hu2020monalog}
Hai Hu, Qi~Chen, Kyle Richardson, Atreyee Mukherjee, Lawrence~S Moss, and
  Sandra Kuebler. 2020.
\newblock Monalog: a lightweight system for natural language inference based on
  monotonicity.

\bibitem[{Hu and Moss(2018)}]{hu2018polarity}
Hai Hu and Larry Moss. 2018.
\newblock Polarity computations in flexible categorial grammar.
\newblock In \emph{Proceedings of the Seventh Joint Conference on Lexical and
  Computational Semantics}, pages 124--129.

\bibitem[{Icard(2012)}]{icard2012inclusion}
Thomas~F Icard. 2012.
\newblock Inclusion and exclusion in natural language.
\newblock \emph{Studia Logica}.

\bibitem[{Icard and Moss(2014)}]{icard2014recent}
Thomas~F Icard and Lawrence~S Moss. 2014.
\newblock Recent progress on monotonicity.
\newblock In \emph{Linguistic Issues in Language Technology}. Citeseer.

\bibitem[{Iftene and Balahur-Dobrescu(2007)}]{Iftene:W07-1421}
Adrian Iftene and Alexandra Balahur-Dobrescu. 2007.
\newblock \emph{Proceedings of the ACL-PASCAL Workshop on Textual Entailment
  and Paraphrasing}. Prague, Czech.

\bibitem[{Iyyer et~al.(2018)Iyyer, Wieting, Gimpel, and Zettlemoyer}]{SCPN}
Mohit Iyyer, John Wieting, Kevin Gimpel, and Luke Zettlemoyer. 2018.
\newblock Adversarial example generation with syntactically controlled
  paraphrase networks.
\newblock In \emph{Proceedings of the 2018 Conference of the North American
  Chapter of the Association for Computational Linguistics: Human Language
  Technologies, Volume 1 (Long Papers)}, pages 1875--1885.

\bibitem[{Jia and Liang(2017)}]{jia2017adversarial}
Robin Jia and Percy Liang. 2017.
\newblock \href {https://doi.org/10.18653/v1/D17-1215} {Adversarial examples
  for evaluating reading comprehension systems}.
\newblock In \emph{Proceedings of the 2017 Conference on Empirical Methods in
  Natural Language Processing (EMNLP)}, pages 2021--2031, Copenhagen, Denmark.
  Association for Computational Linguistics.

\bibitem[{Jin et~al.(2020)Jin, Jin, Zhou, and Szolovits}]{textfooler}
Di~Jin, Zhijing Jin, Joey~Tianyi Zhou, and Peter Szolovits. 2020.
\newblock Is bert really robust? a strong baseline for natural language attack
  on text classification and entailment.
\newblock In \emph{Proceedings of the AAAI conference on artificial
  intelligence}, volume~34, pages 8018--8025.

\bibitem[{Kang et~al.(2018)Kang, Khot, Sabharwal, and Hovy}]{kang18acl}
Dongyeop Kang, Tushar Khot, Ashish Sabharwal, and Eduard Hovy. 2018.
\newblock Adventure: Adversarial training for textual entailment with
  knowledge-guided examples.
\newblock In \emph{The 56th Annual Meeting of the Association for Computational
  Linguistics (ACL)}, Melbourne, Australia.

\bibitem[{Kurakin et~al.(2018)Kurakin, Goodfellow, and
  Bengio}]{kurakin2016adversarial}
Alexey Kurakin, Ian~J Goodfellow, and Samy Bengio. 2018.
\newblock Adversarial examples in the physical world.
\newblock In \emph{Artificial intelligence safety and security}, pages 99--112.
  Chapman and Hall/CRC.

\bibitem[{Lakoff(1970)}]{lakoff1970linguistics}
George Lakoff. 1970.
\newblock Linguistics and natural logic.
\newblock \emph{Synthese}, 22(1-2):151--271.

\bibitem[{Li et~al.(2021)Li, Zhang, Peng, Chen, Brockett, Sun, and
  Dolan}]{li2021clare}
Dianqi Li, Yizhe Zhang, Hao Peng, Liqun Chen, Chris Brockett, Ming-Ting Sun,
  and Bill Dolan. 2021.
\newblock Contextualized perturbation for textual adversarial attack.
\newblock In \emph{Proceedings of the Conference of the North American Chapter
  of the Association for Computational Linguistics (NAACL)}.

\bibitem[{Li et~al.(2020)Li, Ma, Guo, Xue, and Qiu}]{li2020bertattack}
Linyang Li, Ruotian Ma, Qipeng Guo, Xiangyang Xue, and Xipeng Qiu. 2020.
\newblock \href {https://doi.org/10.18653/v1/2020.emnlp-main.500}
  {{BERT}-{ATTACK}: Adversarial attack against {BERT} using {BERT}}.
\newblock In \emph{Proceedings of the 2020 Conference on Empirical Methods in
  Natural Language Processing (EMNLP)}, pages 6193--6202, Online. Association
  for Computational Linguistics.

\bibitem[{Liang et~al.(2018)Liang, Li, Su, Bian, Li, and Shi}]{liang2017deep}
Bin Liang, Hongcheng Li, Miaoqiang Su, Pan Bian, Xirong Li, and Wenchang Shi.
  2018.
\newblock Deep text classification can be fooled.
\newblock In \emph{Proceedings of the 27th International Joint Conference on
  Artificial Intelligence (IJCAI)}, pages 4208--4215.

\bibitem[{Liu et~al.(2019)Liu, Ott, Goyal, Du, Joshi, Chen, Levy, Lewis,
  Zettlemoyer, and Stoyanov}]{roberta}
Yinhan Liu, Myle Ott, Naman Goyal, Jingfei Du, Mandar Joshi, Danqi Chen, Omer
  Levy, Mike Lewis, Luke Zettlemoyer, and Veselin Stoyanov. 2019.
\newblock Roberta: A robustly optimized bert pretraining approach.
\newblock \emph{arXiv preprint arXiv:1907.11692}.

\bibitem[{MacCartney(2009)}]{maccartney2009natural}
Bill MacCartney. 2009.
\newblock \emph{Natural Language Inference}.
\newblock Ph.D. thesis, Stanford University.

\bibitem[{MacCartney and Manning(2009)}]{maccartney2009extend}
Bill MacCartney and Christopher~D Manning. 2009.
\newblock An extended model of natural logic.
\newblock In \emph{Proceedings of the 8th international conference on
  computational semantics (IWCS)}, Stroudsburg, United States.

\bibitem[{Mao et~al.(2019)Mao, Gan, Kohli, Tenenbaum, and
  Wu}]{Mao2019NeuroSymbolic}
Jiayuan Mao, Chuang Gan, Pushmeet Kohli, Joshua~B. Tenenbaum, and Jiajun Wu.
  2019.
\newblock {The Neuro-Symbolic Concept Learner: Interpreting Scenes, Words, and
  Sentences From Natural Supervision}.
\newblock In \emph{Proceedings of the 7th International Conference on Learning
  Representations (ICLR)}, New Orleans, USA.

\bibitem[{Marelli et~al.(2014)Marelli, Menini, Baroni, Bentivogli, Bernardi,
  Zamparelli et~al.}]{sick}
Marco Marelli, Stefano Menini, Marco Baroni, Luisa Bentivogli, Raffaella
  Bernardi, Roberto Zamparelli, et~al. 2014.
\newblock A sick cure for the evaluation of compositional distributional
  semantic models.
\newblock In \emph{Proceedings of the 9th International Conference on Language
  Resources and Evaluation (LREC)}, Reykjavik, Iceland.

\bibitem[{McCoy et~al.(2019)McCoy, Pavlick, and Linzen}]{mccoy2019right}
Tom McCoy, Ellie Pavlick, and Tal Linzen. 2019.
\newblock Right for the wrong reasons: Diagnosing syntactic heuristics in
  natural language inference.
\newblock In \emph{Proceedings of the 57th Annual Meeting of the Association
  for Computational Linguistics (ACL)}, pages 3428--3448.

\bibitem[{Miller(1995)}]{miller1995wordnet}
George~A Miller. 1995.
\newblock Wordnet: a lexical database for english.
\newblock \emph{Communications of the ACM}, 38(11):39--41.

\bibitem[{Minervini and Riedel(2018)}]{minervini2018adversarially}
Pasquale Minervini and Sebastian Riedel. 2018.
\newblock Adversarially regularising neural nli models to integrate logical
  background knowledge.
\newblock In \emph{Proceedings of the 22nd Conference on Computational Natural
  Language Learning (CoNLL)}, pages 65--74.

\bibitem[{Morris et~al.(2020)Morris, Lifland, Yoo, Grigsby, Jin, and
  Qi}]{morris2020textattack}
John Morris, Eli Lifland, Jin~Yong Yoo, Jake Grigsby, Di~Jin, and Yanjun Qi.
  2020.
\newblock Textattack: A framework for adversarial attacks, data augmentation,
  and adversarial training in nlp.
\newblock In \emph{Proceedings of the 2020 Conference on Empirical Methods in
  Natural Language Processing: System Demonstrations}, pages 119--126.

\bibitem[{Nairn et~al.(2006)Nairn, Condoravdi, and
  Karttunen}]{nairn2006computing}
Rowan Nairn, Cleo Condoravdi, and Lauri Karttunen. 2006.
\newblock Computing relative polarity for textual inference.
\newblock In \emph{Proceedings of the 5th international workshop on inference
  in computational semantics}, Buxton, England.

\bibitem[{Partee(1995)}]{partee1995}
Barbara Partee. 1995.
\newblock Lexical semantics and compositionality.
\newblock \emph{Invitation to Cognitive Science}.

\bibitem[{Pilault et~al.(2021)Pilault, Pal et~al.}]{pilault2021conditionally}
Jonathan Pilault, Christopher Pal, et~al. 2021.
\newblock Conditionally adaptive multi-task learning: Improving transfer
  learning in nlp using fewer parameters \& less data.
\newblock In \emph{International Conference on Learning Representations
  (ICLR)}.

\bibitem[{Poliak et~al.(2018)Poliak, Haldar, Rudinger, Hu, Pavlick, White, and
  Van~Durme}]{poliak2018collecting}
Adam Poliak, Aparajita Haldar, Rachel Rudinger, J.~Edward Hu, Ellie Pavlick,
  Aaron~Steven White, and Benjamin Van~Durme. 2018.
\newblock \href {https://doi.org/10.18653/v1/D18-1007} {Collecting diverse
  natural language inference problems for sentence representation evaluation}.
\newblock In \emph{Proceedings of the 2018 Conference on Empirical Methods in
  Natural Language Processing}, pages 67--81, Brussels, Belgium. Association
  for Computational Linguistics.

\bibitem[{Radford et~al.(2019)Radford, Wu, Child, Luan, Amodei, and
  Sutskever}]{gpt2}
Alec Radford, Jeffrey Wu, Rewon Child, David Luan, Dario Amodei, and Ilya
  Sutskever. 2019.
\newblock Language models are unsupervised multitask learners.
\newblock \emph{OpenAI Blog}, 1(8):9.

\bibitem[{Ren et~al.(2019)Ren, Deng, He, and Che}]{pwws}
Shuhuai Ren, Yihe Deng, Kun He, and Wanxiang Che. 2019.
\newblock Generating natural language adversarial examples through probability
  weighted word saliency.
\newblock In \emph{Proceedings of the 57th Annual Meeting of the Association
  for Computational Linguistics (ACL)}, pages 1085--1097.

\bibitem[{Richardson et~al.(2020)Richardson, Hu, Moss, and
  Sabharwal}]{fragments2020}
Kyle Richardson, Hai Hu, Lawrence~S Moss, and Ashish Sabharwal. 2020.
\newblock Probing natural language inference models through semantic fragments.
\newblock In \emph{Proceedings of the Thirty-Fourth AAAI Conference on
  Artificial Intelligence (AAAI)}, New York, USA.

\bibitem[{Rockt{\"a}schel and Riedel(2017)}]{rocktaschel2017end}
Tim Rockt{\"a}schel and Sebastian Riedel. 2017.
\newblock End-to-end differentiable proving.
\newblock In \emph{Proceedings of the 31st International Conference on Neural
  Information Processing Systems (NeurIPS)}, Long Beach, USA.

\bibitem[{Salazar et~al.(2020)Salazar, Liang, Nguyen, and
  Kirchhoff}]{salazar2020masked}
Julian Salazar, Davis Liang, Toan~Q. Nguyen, and Katrin Kirchhoff. 2020.
\newblock \href {https://doi.org/10.18653/v1/2020.acl-main.240} {Masked
  language model scoring}.
\newblock In \emph{Proceedings of the 58th Annual Meeting of the Association
  for Computational Linguistics}, pages 2699--2712, Online. Association for
  Computational Linguistics.

\bibitem[{Sanh et~al.(2019)Sanh, Debut, Chaumond, and
  Wolf}]{sanh2019distilbert}
Victor Sanh, Lysandre Debut, Julien Chaumond, and Thomas Wolf. 2019.
\newblock Distilbert, a distilled version of bert: smaller, faster, cheaper and
  lighter.
\newblock \emph{the 5th Workshop on Energy Efficient Machine Learning and
  Cognitive Computing @NeurIPS}.

\bibitem[{Sato et~al.(2018)Sato, Suzuki, Shindo, and
  Matsumoto}]{sato2018interpretable}
Motoki Sato, Jun Suzuki, Hiroyuki Shindo, and Yuji Matsumoto. 2018.
\newblock Interpretable adversarial perturbation in input embedding space for
  text.
\newblock In \emph{Proceedings of the 27th International Joint Conference on
  Artificial Intelligence (IJCAI)}, pages 4323--4330.

\bibitem[{Tram{\`e}r et~al.(2020)Tram{\`e}r, Behrmann, Carlini, Papernot, and
  Jacobsen}]{tramer2020fundamental}
Florian Tram{\`e}r, Jens Behrmann, Nicholas Carlini, Nicolas Papernot, and
  J{\"o}rn-Henrik Jacobsen. 2020.
\newblock Fundamental tradeoffs between invariance and sensitivity to
  adversarial perturbations.
\newblock In \emph{International Conference on Machine Learning}, pages
  9561--9571. PMLR.

\bibitem[{Valencia(1991)}]{valencia1991studies}
V{\'\i}ctor Manuel~S{\'a}nchez Valencia. 1991.
\newblock \emph{Studies on natural logic and categorial grammar}.
\newblock Universiteit van Amsterdam.

\bibitem[{van Benthem(1988)}]{van1988semantics}
Johan van Benthem. 1988.
\newblock The semantics of variety in categorial grammar.
\newblock \emph{Categorial grammar}.

\bibitem[{Van~Benthem(1995)}]{van1995language}
Johan Van~Benthem. 1995.
\newblock \emph{Language in Action: categories, lambdas and dynamic logic}.
\newblock MIT Press.

\bibitem[{Van~Benthem et~al.(1986)}]{van1986essays}
Johan Van~Benthem et~al. 1986.
\newblock \emph{Essays in logical semantics}.
\newblock Springer.

\bibitem[{Wallace et~al.(2019)Wallace, Feng, Kandpal, Gardner, and
  Singh}]{wallace2019universal}
Eric Wallace, Shi Feng, Nikhil Kandpal, Matt Gardner, and Sameer Singh. 2019.
\newblock \href {https://doi.org/10.18653/v1/D19-1221} {Universal adversarial
  triggers for attacking and analyzing {NLP}}.
\newblock In \emph{Proceedings of the 2019 Conference on Empirical Methods in
  Natural Language Processing and the 9th International Joint Conference on
  Natural Language Processing (EMNLP-IJCNLP)}, pages 2153--2162, Hong Kong,
  China. Association for Computational Linguistics.

\bibitem[{Weber et~al.(2019)Weber, Minervini, M{\"u}nchmeyer, Leser, and
  Rockt{\"a}schel}]{weber2019nlprolog}
Leon Weber, Pasquale Minervini, Jannes M{\"u}nchmeyer, Ulf Leser, and Tim
  Rockt{\"a}schel. 2019.
\newblock Nlprolog: Reasoning with weak unification for question answering in
  natural language.
\newblock In \emph{Proceedings of the 57th Annual Meeting of the Association
  for Computational Linguistics (ACL)}, Austin, Texas, United States.

\bibitem[{Williams et~al.(2018)Williams, Nangia, and Bowman}]{mnli}
Adina Williams, Nikita Nangia, and Samuel Bowman. 2018.
\newblock \href {https://doi.org/10.18653/v1/N18-1101} {A broad-coverage
  challenge corpus for sentence understanding through inference}.
\newblock In \emph{Proceedings of the 2018 Conference of the North {A}merican
  Chapter of the Association for Computational Linguistics: Human Language
  Technologies, Volume 1 (Long Papers)}, pages 1112--1122, New Orleans,
  Louisiana. Association for Computational Linguistics.

\bibitem[{Yanaka et~al.(2020)Yanaka, Mineshima, Bekki, and
  Inui}]{systematicity2020}
Hitomi Yanaka, Koji Mineshima, Daisuke Bekki, and Kentaro Inui. 2020.
\newblock \href {https://doi.org/10.18653/v1/2020.acl-main.543} {Do neural
  models learn systematicity of monotonicity inference in natural language?}
\newblock In \emph{Proceedings of the 58th Annual Meeting of the Association
  for Computational Linguistics (ACL)}, pages 6105--6117.

\bibitem[{Yanaka et~al.(2019{\natexlab{a}})Yanaka, Mineshima, Bekki, Inui,
  Sekine, Abzianidze, and Bos}]{med2019}
Hitomi Yanaka, Koji Mineshima, Daisuke Bekki, Kentaro Inui, Satoshi Sekine,
  Lasha Abzianidze, and Johan Bos. 2019{\natexlab{a}}.
\newblock Can neural networks understand monotonicity reasoning?
\newblock In \emph{Proceedings of the 2019 ACL Workshop BlackboxNLP: Analyzing
  and Interpreting Neural Networks for NLP}, Austin, Texas, United States.

\bibitem[{Yanaka et~al.(2019{\natexlab{b}})Yanaka, Mineshima, Bekki, Inui,
  Sekine, Abzianidze, and Bos}]{help2019}
Hitomi Yanaka, Koji Mineshima, Daisuke Bekki, Kentaro Inui, Satoshi Sekine,
  Lasha Abzianidze, and Johan Bos. 2019{\natexlab{b}}.
\newblock Help: A dataset for identifying shortcomings of neural models in
  monotonicity reasoning.
\newblock In \emph{Proceedings of the Eighth Joint Conference on Lexical and
  Computational Semantics (*SEM)}, Minneapolis, Minnesota, USA.

\bibitem[{Yang et~al.(2017)Yang, Yang, and Cohen}]{yang2017differentiable}
Fan Yang, Zhilin Yang, and William~W Cohen. 2017.
\newblock Differentiable learning of logical rules for knowledge base
  reasoning.
\newblock In \emph{Advances in Neural Information Processing Systems}, pages
  2319--2328.

\bibitem[{Zang et~al.(2020)Zang, Qi, Yang, Liu, Zhang, Liu, and Sun}]{pso}
Yuan Zang, Fanchao Qi, Chenghao Yang, Zhiyuan Liu, Meng Zhang, Qun Liu, and
  Maosong Sun. 2020.
\newblock Word-level textual adversarial attacking as combinatorial
  optimization.
\newblock In \emph{Proceedings of the 58th Annual Meeting of the Association
  for Computational Linguistics (ACL)}, pages 6066--6080.

\bibitem[{Zeng et~al.(2021)Zeng, Qi, Zhou, Zhang, Hou, Zang, Liu, and
  Sun}]{zeng2020openattack}
Guoyang Zeng, Fanchao Qi, Qianrui Zhou, Tingji Zhang, Bairu Hou, Yuan Zang,
  Zhiyuan Liu, and Maosong Sun. 2021.
\newblock \href {https://doi.org/10.18653/v1/2021.acl-demo.43} {{Openattack: An
  open-source textual adversarial attack toolkit}}.
\newblock In \emph{Proceedings of the 59th Annual Meeting of the Association
  for Computational Linguistics and the 11th International Joint Conference on
  Natural Language Processing: System Demonstrations}, pages 363--371.

\bibitem[{Zhang et~al.(2019)Zhang, Zhou, Miao, and Li}]{zhang2020generating}
Huangzhao Zhang, Hao Zhou, Ning Miao, and Lei Li. 2019.
\newblock \href {https://doi.org/10.18653/v1/P19-1559} {Generating fluent
  adversarial examples for natural languages}.
\newblock In \emph{Proceedings of the 57th Annual Meeting of the Association
  for Computational Linguistics (ACL)}, pages 5564--5569, Florence, Italy.
  Association for Computational Linguistics.

\bibitem[{Zhang et~al.(2020)Zhang, Wu, Zhao, Li, Zhang, Zhou, and
  Zhou}]{zhang2020semantics}
Zhuosheng Zhang, Yuwei Wu, Hai Zhao, Zuchao Li, Shuailiang Zhang, Xi~Zhou, and
  Xiang Zhou. 2020.
\newblock Semantics-aware bert for language understanding.
\newblock In \emph{Proceedings of the AAAI Conference on Artificial
  Intelligence}, volume~34, pages 9628--9635.

\bibitem[{Zhao et~al.(2018)Zhao, Dua, and Singh}]{zhao2017generating}
Zhengli Zhao, Dheeru Dua, and Sameer Singh. 2018.
\newblock Generating natural adversarial examples.
\newblock In \emph{Proceedings of the 6th International Conference on Learning
  Representations (ICLR)}.

\end{thebibliography}
\bibliographystyle{acl_natbib}

\appendix
\begin{table*}
\definecolor{lightgray}{gray}{0.90}
\begin{minipage}[!t]{0.68\linewidth}
\resizebox{1\linewidth}{!}{
\setlength{\tabcolsep}{5pt}
\begin{tabular}{cccc}
\toprule
\textbf{Relation} & \textbf{Relation Name} & \textbf{Example} & \textbf{Set Theoretic Definition}  \\
\midrule
  $x \equiv y$  & equivalence &  $mommy \equiv mother$  & $x = y$  \\
  $x \sqsubset y$  & forward entailment &  $bird \sqsubset animal$ &  $x \subset y$ \\
  $x \sqsupset y$  & reverse entailment &  $animal \sqsupset bird$  &  $x \supset y$  \\
  $x$ \textsuperscript{$\wedge$} $y$  & negation & $human$ \textsuperscript{$\wedge$} $ nonhuman$   & $x \cap y = \varnothing \wedge x \cup y = U$  \\
  $x \mid y$  & alternation & $bird \mid dog$   & $x \cap y = \varnothing \wedge x \cup y \ne U$   \\
  $x \smallsmile y$  & cover & $ animal  \smallsmile nonhuman $   & $x \cap y \ne  \varnothing  \wedge x \cup y = U$   \\
  $x \ \# \ y$  & independence & $ red \   \#  \ weak$   & all other cases   \\
\bottomrule
\end{tabular}
}
\caption{Seven natural logic relations proposed by~\protect\citet{maccartney2009extend}.} 
\label{table:basic-rel}

\end{minipage}
\hfill
\begin{minipage}[!t]{0.3\linewidth}

\resizebox{1\linewidth}{!}{
\begin{tabular}{|>{\columncolor{lightgray}}c|c|c|c|c|c|c|c|}
\hline
\rowcolor{lightgray}
$\Join$ & $\equiv$  &   $\sqsubset$ &   $\sqsupset$ &   $\wedge$  & $\mid$    &   $\smallsmile$   &    $\#$  \\
\hline
$\equiv$ & $\equiv$  &   $\sqsubset$ &   $\sqsupset$ &   $\wedge$  & $\mid$    &   $\smallsmile$   &    $\#$  \\
$\sqsubset$ & $\sqsubset$  &   $\sqsubset$ &   $\#$ &   $\mid$  & $\mid$    &   $\#$   &    $\#$  \\

$\sqsupset$ & $\sqsupset$  &   $\#$ &   $\sqsupset$ &   $\smallsmile$  & $\#$    &   $\smallsmile$   &    $\#$  \\

$\wedge$ &  $\wedge$  &   $\smallsmile$ &   $\mid$ &   $\equiv$  & $\sqsupset$    &   $\sqsubset$   &    $\#$  \\

$\mid$ &  $\mid$  &   $\#$ &   $\mid$ &   $\sqsubset$  & $\#$    &   $\sqsubset$   &    $\#$  \\

$\smallsmile$ & $\smallsmile$  &   $\smallsmile$ &   $\#$ &  $\sqsupset$ &   $\sqsupset$  & $\#$    &    $\#$    \\

$\#$ &   $\#$&   $\#$&   $\#$&   $\#$&   $\#$&   $\#$&   $\#$ \\
\hline
\end{tabular}
}
\caption{Relation union table~\cite{icard2012inclusion}. 
}
\label{table:union}
\end{minipage}
\end{table*}

\section{Background}
\label{sec:background}

Our work is based on the specific natural logic formalism proposed by \citet{maccartney2009extend} and \citet{maccartney2009natural}.
To model the entailment relations between two spans of texts, \citet{maccartney2009extend} introduced seven relations inspired by the set theory: $\frak{B}=\{\,\equiv, \sqsubset, \sqsupset, \wedge, \,\mid\,,  \smallsmile,\allowbreak \#\,\}$   (see Table~\ref{table:basic-rel} for some examples). 
The inference of natural logic is built on monotonicity, which is a pervasive feature of natural language that explains the impact of semantic composition on entailment relations~\cite{van1986essays,valencia1991studies,icard2014recent}.  Suppose \textit{dog} $\sqsubset$ \textit{animal},  
the upward monotone context keeps the entailment relation when the argument “increases” (e.g., \textit{dog} $\sqsubset$ \textit{animal}). Downward monotone keeps the entailment relation when the argument “decreases” (e.g., in \textit{all animals} $\sqsubset$ \textit{all dogs}). The system performs monotonicity inference through a projection $\rho \colon \mathfrak{B} \rightarrow \mathfrak{B}$, which is determined by the context and projection rules. As will be detailed, a monotonicity-based parser can provide monotonicity information for each word token in a sentence and the projection information. For example, consider the sentence \textit{All$\uparrow$ the$\downarrow$ kids$\downarrow$ run$\uparrow$}, where $\uparrow$ denoted upward polarity and $\downarrow$ downward polarity. If we mutate the word \textit{kids} with \textit{boys}, where \textit{kids} $\sqsupset$ \textit{boys}, the system projects the \textit{reverse entailment}~(`$\sqsupset$') into \textit{forward entailment}~(`$\sqsubset$') due to its downward polarity, \ie $\rho$ (`$\sqsupset$')  = `$\sqsubset$', and thus \textit{All the kids run} $\sqsubset$ \textit{All the boys run}.

With these components ready, the system aggregates the projected local relations to obtain the inferential relation between a premise and hypothesis sentence. Specifically, Table~\ref{table:union} \cite{maccartney2009natural,maccartney2009extend, angeli2014naturalli} shows the composition function when a relation in the first column is joined with a relation listed in the first row, yielding the relations in the corresponding table cell.
\citet{maccartney2009natural} shows that different orders of compositions yield consistent results except in some rare artificial cases. Therefore, many works, including ours, perform a sequential (left-to-right) composition. 
Consider two edits from the premise sentence, \textit{All the kids run}, to the hypothesis, \textit{All the boys sleep}. The first edit that replaces \textit{kids} in the premise with \textit{boys} yields \textit{All the kids run} $\sqsubset$ \textit{All the boys run}. The second edit of replacing \textit{run} with \textit{sleep} yields \textit{All the boys run} $\mid$ \textit{All the boys sleep}. Based on Table~\ref{table:union}, the union of the relations resulted from these two edits (i.e., `$\sqsubset$' $\Join$ `$\mid$') is `$\,\mid\,$', where $\Join$ is the union operator. As a result, we obtain \textit{All the kids run} $\,\mid\,$ \textit{All the boys sleep}. 

The seven natural logic relations at the sentence-pair level can then be mapped to the typical three-way NLI labels~(\textit{entailment}, \textit{contradiction}, and \textit{neutral}), where the `$\,\equiv\,$' or `$\,\sqsubset\,$' relation can be mapped to \textit{entailment}; the `$\,\wedge\,$' or `\,$\mid$\,' relation to \textit{contradiction}; the `$\,\sqsupset\,$', `$\,\smallsmile\,$', and `$\,\#\,$' to \textit{neutral}. 

\section{Insertion and Deletion}
\label{sec:insertion}
For both insertion and deletion, the part-of-speech~(POS) tags and constituency parse tree for $H^{(1)}$ are first obtained using Stanford \textit{CoreNLP} parser\footnote{https://stanfordnlp.github.io}, which are then used with a state-of-the-art pretrained model to perform insertion. To insert an \textit{adjective} before a \textit{noun} or an \textit{adverb} after a \textit{verb}, \texttt{NatLogAttack} leverages DistilBert~\cite{sanh2019distilbert} to obtain the candidates in the corresponding locations.  The syntactic rules do not guarantee to generate sentences with the desired NLI labels (e.g., see \cite{partee1995} for discussion on the semantic composition of \textit{adjective}\ +\ \textit{noun}). The above process is only for generating candidates, and we will use pretrained language models to find good adversarial examples.

In order to insert a prepositional phrase~(PP), we first collected from the SNLI training dataset all the PPs that are the constitutes of other noun phrases (NPs) for more than 100 times.
We also collected PPs that appear in other verb phrases (VPs) at least 100 times.
During insertion, these PPs will be added as modifiers to a noun or a verb, respectively. We also insert assertion phrases such as "It is not true that" to deceive the victim models.
For the \textit{deletion} operation, we delete the corresponding constituents based on the parse tree and POS tags.

\section{Details of Datasets and Baselines}
\label{sec:baselineDetails}
As discussed in Section~\ref{sec:expsetup}, our study uses SNLI~\cite{snli}, MNLI~\cite{mnli}, MED~\cite{med2019}, HELP~\cite{help2019}, and SICK~\cite{sick,hu2020monalog} to evaluate the models.
SNLI and MNLI are widely-used general-purpose NLI datasets. Following~\citet{li2021clare}, for MNLI, we evaluate the performance on the \textit{matched} set. MED and HELP are designed for monotonicity-based reasoning and hence suit for probing models' capacity in natural logic-related behaviour. SICK is rich in lexical, syntactic and semantic phenomena designed for distributional semantic models including those recognizing textual entailment. For SICK, we use the corrected labels proposed by~\citet{hu2020monalog}.    
The pretrained victim models tested on the SNLI, MNLI, and SICK test set were finetuned on their own training set and the performances are comparable to the state-of-the-art performances as well as those used in the previous attack models. Following \citet{med2019}, the models tested on MED are finetuned on both the SNLI training set and the entire HELP dataset. Since HELP is not manually annotated, we do not use it as the test set.
The MED upward subset is denoted as $\text{MED}_{\text{up}}$ and downward subset as $\text{MED}_{\text{down}}$. 
Following ~\cite{ga,pso}, each test set has 1,000 sentence pairs. Also following ~\citet{zeng2020openattack}, we set the maximum query number to be 500. 

For all the attack models in comparison, we used the implementation made available by \citet{morris2020textattack}.
Details of these attack models are as follows.
\begin{itemize}
     

    \item\textbf{PWWS}~\cite{pwws} makes use of the synonyms in WordNet~\cite{miller1995wordnet} for word substitutions and designs a greedy search algorithm based on the probability-weighted word saliency to generate adversarial samples. 
    
    \item\textbf{TextFooler}~\cite{textfooler} utilizes counter-fitting word embeddings to obtain synonyms and then performs substitution based on that.

    \item\textbf{PSO}~\cite{pso} utilizes the knowledge base HowNet~\cite{dong2006hownet} to generate word substitutions. It adopts particle swarm optimization, another popular meta-heuristic population-based search algorithm, as its search strategy.

    \item\textbf{BertAttack}~\cite{li2020bertattack} leverages the superior performance of pretrained language model and greedily replaces tokens with the predictions from BERT.

    \item\textbf{Clare}~\cite{li2021clare} adds two more types of perturbations, \textit{insert} and \textit{merge}, building on BertAttack. Since Clare has a very high query number to the victim models, we reduce the number of each type of perturbation to 10 in order to make sure that Clare can attack the victim model successfully within the maximum query number in most cases. 

\end{itemize}

\end{document}